\RequirePackage{fix-cm}
\documentclass[twocolumn]{svjour3}          % twocolumn
\smartqed  % flush right qed marks, e.g. at end of proof
\usepackage{graphicx}
\usepackage{multirow}
\usepackage{amsmath,amssymb,amsfonts}
\usepackage{mathrsfs}
\usepackage[dvipsnames,svgnames,x11names,table]{xcolor}
\usepackage{textcomp}
\usepackage{booktabs}
\usepackage{pifont}
\usepackage{url}
\usepackage[ruled,vlined,linesnumbered]{algorithm2e}
\usepackage{listings}
\usepackage{xspace}
\usepackage{threeparttable}
\usepackage{adjustbox}

\usepackage[utf8]{inputenc}
\usepackage[T1]{fontenc}
\usepackage{url}
\usepackage{nicefrac}
\usepackage{microtype}
\usepackage{epsfig}
\usepackage{float}
\usepackage{multicol}
\usepackage{wrapfig}
\usepackage[misc]{ifsym}
\usepackage{makecell}
\usepackage{soul}
\usepackage{bm}
\usepackage{subcaption, overpic, textpos}

\usepackage{ulem}
\usepackage{pifont}
\usepackage{caption}
\usepackage{array}
\usepackage{enumitem}
\usepackage{cite}
\usepackage{tablefootnote}
\usepackage{diagbox}
\usepackage{tabularx}

\newcolumntype{C}[1]{>{\centering\arraybackslash}p{#1}}

\definecolor{linkcolor}{RGB}{255,0,0}
\definecolor{urlcolor}{RGB}{255,105,180}
\definecolor{citecolor}{RGB}{66,168,235}
\usepackage[pagebackref,breaklinks=true,colorlinks,citecolor=blue,urlcolor=blue,linkcolor=blue,bookmarks=false]{hyperref}

\usepackage[most]{tcolorbox}
\usepackage{tabularx}
\usepackage{xcolor}
\usepackage{pifont}

\definecolor{schemaAnchor}{HTML}{F3B23C}
\definecolor{schemaVisual}{HTML}{4C9A64}
\definecolor{schemaAudio}{HTML}{7E6BB7}
\definecolor{schemaShot}{HTML}{E08A2E}
\definecolor{schemaBind}{HTML}{D95C75}
\definecolor{schemaGray}{HTML}{F7F7F7}

\tcbset{
  schemaBox/.style={
    enhanced,
    colback=#1!7,
    colframe=#1!75!black,
    boxrule=0.6pt,
    arc=2mm,
    left=4pt,
    right=4pt,
    top=3pt,
    bottom=3pt,
    fonttitle=\bfseries\small,
    coltitle=black,
    before skip=3pt,
    after skip=3pt
  }
}

\newlength\savewidth

\renewcommand{\paragraph}[1]{\vspace{1.25mm}\noindent\textbf{#1}}

\definecolor{lightgray}{rgb}{0.8, 0.8, 0.8}
\definecolor{lgray}{rgb}{0.66, 0.66, 0.66}
\definecolor{whit_tab}{RGB}{255, 255, 255}
\definecolor{gray_tab}{RGB}{235, 235, 235}
\definecolor{oran_tab}{RGB}{254, 247, 241}
\definecolor{blue_tab}{RGB}{200, 227, 245}
\definecolor{lblu_tab}{RGB}{231, 239, 248}
\definecolor{teaser_blue}{RGB}{106, 153, 208}
\definecolor{teaser_orange}{RGB}{222, 131, 68}
\definecolor{TableAccent}{HTML}{E5F0FA}

\definecolor{BenchGreen}{HTML}{16823A}

\definecolor{BenchRed}{HTML}{B51E1E}

\definecolor{BenchBlue}{HTML}{4477AA}

\newcommand{\BenchYes}{\textcolor{BenchGreen}{\ding{51}}}

\newcommand{\BenchNo}{\textcolor{BenchRed}{\ding{55}}}

\newcommand{\BenchPart}{\textcolor{BenchBlue}{\ding{109}}}

\newcommand{\Omark}{\hspace*{2pt}\colorbox{red!10}{\textcolor{red!80!black}{\fontsize{6.5pt}{6.5pt}\selectfont\textbf{\textsf{O}}}}}
\newcommand{\Lmark}{\hspace*{2pt}\colorbox{green!15}{\textcolor{green!50!black}{\fontsize{6.5pt}{6.5pt}\selectfont\textbf{\textsf{L}}}}}

\newcommand{\Ours}{CineDance-1M\xspace}        % dataset name
\newcommand{\OurBench}{CineBench\xspace}        % benchmark name
\newcommand{\OurModel}{CineDance\xspace}        % model name
\newcommand{\AvgDur}{92.8\xspace}               % average duration (seconds)
\newcommand{\AvgShots}{24.2\xspace}             % average shot count
\newcommand{\VideoHours}{31,530\xspace}          % total video hours
\newcommand{\RawFiles}{45,181\xspace}            % raw file count

\begin{document}
\begin{sloppypar}

\title{CineDance: Towards Next-Generation Multi-Shot Long-Form Cinematic Audio-Video Generation}

% \author{Yuheng Chen$^{1,\dagger}$ \and
%         Teng Hu$^{1,\dagger}$ \and
%         Yuji Wang$^{1}$ \and
%         Qingdong He$^{2}$ \and
%         Zhucun Xue$^{3}$ \and
%         Qianyu Zhou$^{4}$ \and
%         Xiangtai Li$^{5}$ \and
%         Lizhuang Ma$^{1,*}$ \and
%         Jiangning Zhang$^{3}$ \and
%         Dacheng Tao$^{5}$
% }

\author{Yuheng Chen$^{1,*}$ \and
        Teng Hu$^{1,*}$ \and
        Yuji Wang$^{1}$ \and
        Qingdong He$^{2}$ \and
        Zhucun Xue$^{3}$ \and
        Qianyu Zhou$^{4}$ \and
        % Xiangtai Li$^{5}$ \and
        Jason Li$^{5}$ \and
        Lizhuang Ma$^{1,\dagger}$ \and
        Jiangning Zhang$^{3,\ddagger}$ \and
        Dacheng Tao$^{5}$
}

% \author{Yuheng Chen$^{1}$ \and
%         Teng Hu$^{1}$ \and
%         Yuji Wang$^{1}$ \and
%         Qingdong He$^{2}$ \and
%         Zhucun Xue$^{3}$ \and
%         Qianyu Zhou$^{4}$ \and
%         Xiangtai Li$^{5}$ \and
%         Lizhuang Ma$^{1}$ \and
%         Jiangning Zhang$^{3}$ \and
%         Dacheng Tao$^{5}$
% }

\authorrunning{Chen et al.}

\institute{
  % $^{\dagger}$Equal contribution. $^{*}$Corresponding author: Lizhuang Ma (lzma@sjtu.edu.cn) \\
  $^{*}$Equal contribution. $^{\dagger}$Corresponding author. $^{\ddagger}$Project lead. \\ 
  $^{1}$ Shanghai Jiao Tong University, Shanghai, China. \\
  $^{2}$ University of Electronic Science and Technology of China, Chengdu, China. \\
  $^{3}$ Zhejiang University, Hangzhou, China. \\
  $^{4}$ The University of Tokyo, Tokyo, Japan. \\
  $^{5}$ Nanyang Technological University, Singapore, Singapore. \\
}

\date{Received: date / Accepted: date}
\maketitle

\begin{abstract}
The fidelity and structural diversity of training datasets fundamentally determine the capabilities of video generation models. While commercial systems show remarkable ability to generate cinematic narratives, the progress of open-source models remains limited by the scarcity of high-quality training data.
To bridge this gap, we introduce \Ours, a large-scale, open research Text-to-Audio-Video (T2AV) dataset designed specifically for multi-shot, long-form joint audio-video generation. Averaging \AvgDur seconds and \AvgShots continuous shots per video, it provides configurable, structured annotations for both audio and video modalities. This exceptional quality is achieved through a rigorous three-stage curation pipeline: \textit{i)} diverse sourcing and comprehensive cleansing, \textit{ii)} film-theory-inspired narrative parsing, and \textit{iii)} hierarchical dual-modal captioning.
For a comprehensive assessment, we propose \OurBench, featuring a diverse prompt suite and a six-dimensional, human-aligned metric system tailored for complex narrative audio-video evaluation.
Furthermore, we adapt LTX-2.3 into \OurModel, which demonstrates exceptional single-modality quality alongside precise audio-video alignment and robust subject and environment consistency, effectively validating our curation strategy and the high quality of \Ours.
We anticipate that this work will serve as a solid foundation for accelerating future research in multi-shot, long-form joint audio-video generation. 
Our project page is available 
% \href{https://aliothchen.github.io/projects/CineDance/}{here}.
at \href{https://aliothchen.github.io/projects/CineDance/}{https://aliothchen.github.io/projects/CineDance/}.
\end{abstract}

% \keywords{video generation, audio-video generation, multi-shot generation, cinematic datasets, video understanding}

\begin{figure*}[htp]
    \centering
    \includegraphics[width=\linewidth]{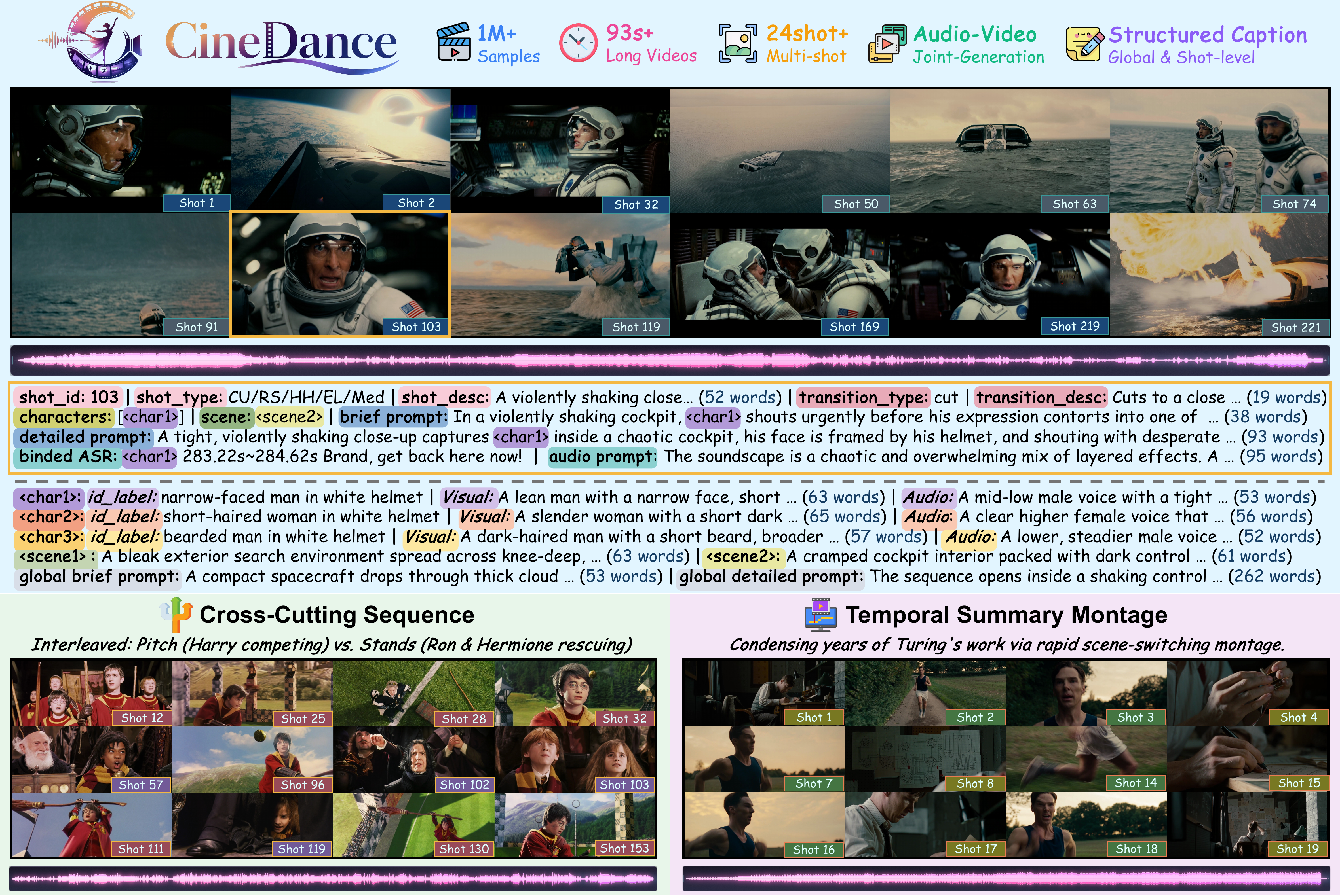}
    \caption{
    \Ours features \textbf{1M} unprecedented long-form (\textbf{\AvgDur}\,s) and multi-shot (\textbf{\AvgShots} shots) \textbf{audio-video} sequences (above), paired with \textbf{hierarchical structured captions} for both modalities. Compared with typical Text-To-Video (T2V) datasets, it encompasses \textbf{diverse narrative structures} (below), meeting the growing demand for cinematic, narrative-driven joint generation.
    }
    \label{fig:teaser}
    \vspace{-6pt}
\end{figure*}

\section{Introduction}
\label{sec:intro}

The unprecedented evolution of video generative models has catalyzed a growing demand for high-fidelity visual content across film production, immersive media, and interactive entertainment~\cite{hu_survey,seedance2,wan,cogvideox,generative_survey,hunyuanvideo,hunyuancustom,ultragen}.
While current works have achieved remarkable visual excellence in generating single-shot clips, the transition to multi-shot, long-form narrative generation remains largely underexplored.
This progression is primarily bottlenecked by the scarcity of such large-scale open-source datasets, coupled with the limited generalizability of existing foundation models to long-form generation, leaving a vast landscape for future exploration.

\begin{figure}[t]
    \centering
    \includegraphics[width=\linewidth]{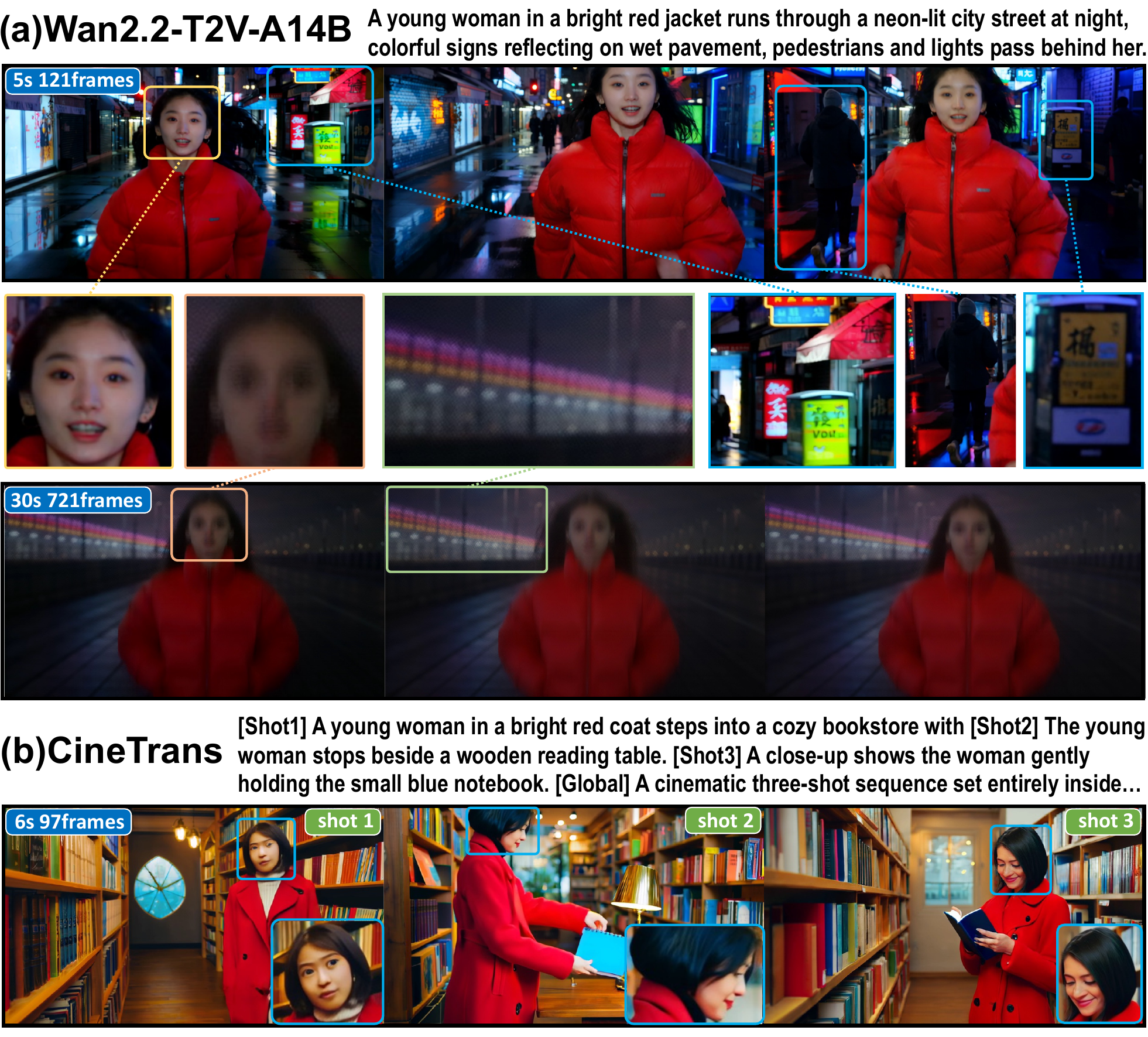}
    \caption{Diagnostic examples illustrating two core challenges in multi-shot long-form generation.
% \textbf{Top:} 
(a) Wan2.2-T2V-A14B produces plausible 5-second results but degrades at 30 seconds, with heavy spatial blurring, low-amplitude motion, and near-static dynamics.
% \textbf{Bottom:} 
(b) CineTrans generates three-shot transitions, but character identity is poorly preserved across shots.
\textit{Shortened prompt summaries are shown for readability.}}
    \label{fig:intro_failure_cases}
\vspace{-10pt}
\end{figure}

% added
However, multi-shot long-form audio-video generation is not a straightforward extension of short-clip synthesis, since it faces \textbf{\textit{two core challenges}}.
Firstly, \textbf{long-horizon scalability}: Foundation models trained and optimized primarily for short clips generally struggle to generalize to extended cinematic durations.
As shown in Fig.~\ref{fig:intro_failure_cases}(a), when Wan2.2-T2V-A14B~\cite{wan} is directly applied to a 30-second continuous-motion prompt, its generation quality noticeably deteriorates compared with the short-form output, exhibiting spatial blurring, reduced motion amplitude, repetitive background patterns, and near-static dynamics.
This suggests that strong short-clip synthesis capability does not necessarily imply stable long-form generation.
Secondly, \textbf{cross-shot semantic consistency}: Multi-shot generation requires characters, objects, and scenes to remain visually consistent and semantically identifiable across discrete shots; joint audio-video generation further demands audio and audio-visual consistency.
Although dedicated multi-shot methods such as CineTrans~\cite{cinetrans} can introduce cinematic shot transitions, they still struggle with cross-shot entity consistency even within relatively short clips, as illustrated in Fig.~\ref{fig:intro_failure_cases}(b). 
For a prompt describing the same character and scene recurring across multiple shots, CineTrans can generate plausible shot transitions, yet may fail to preserve the character’s identity and state, as well as scene consistency. 
% Therefore, long-form cinematic generation requires not only per-shot visual fidelity, but also robust identity preservation, scene consistency, and ordered event execution across shots.

\begin{figure}[t]
    \centering

    \begin{subfigure}[t]{\linewidth}
        \centering
        \includegraphics[width=\linewidth]{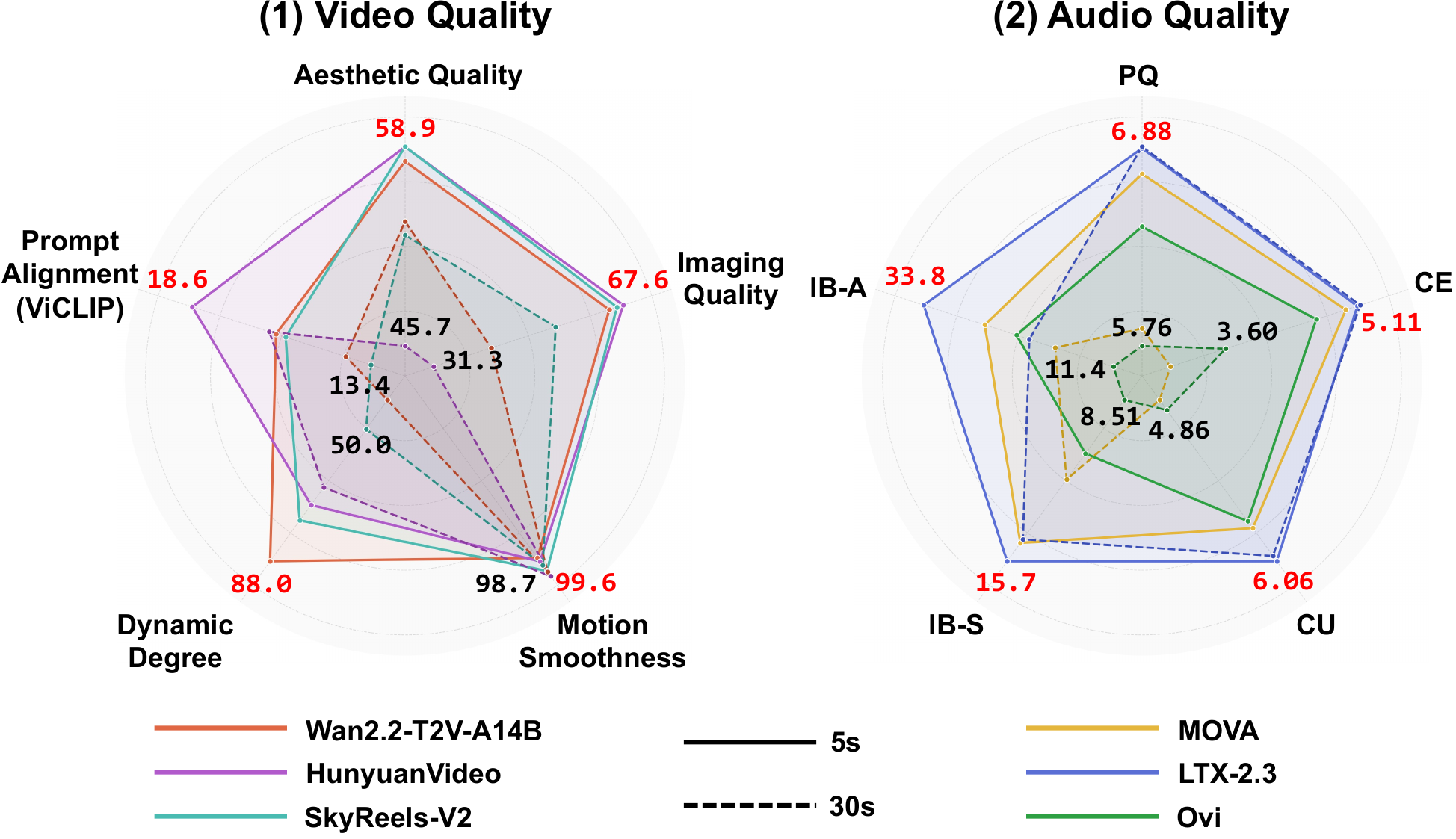}
        \caption{Quality comparison between 5s and 30s generated videos from popular foundation models.}
        \label{fig:radar}
    \end{subfigure}

    % \vspace{0.6em}

    \begin{subfigure}[t]{\linewidth}
        \centering
        \includegraphics[width=\linewidth]{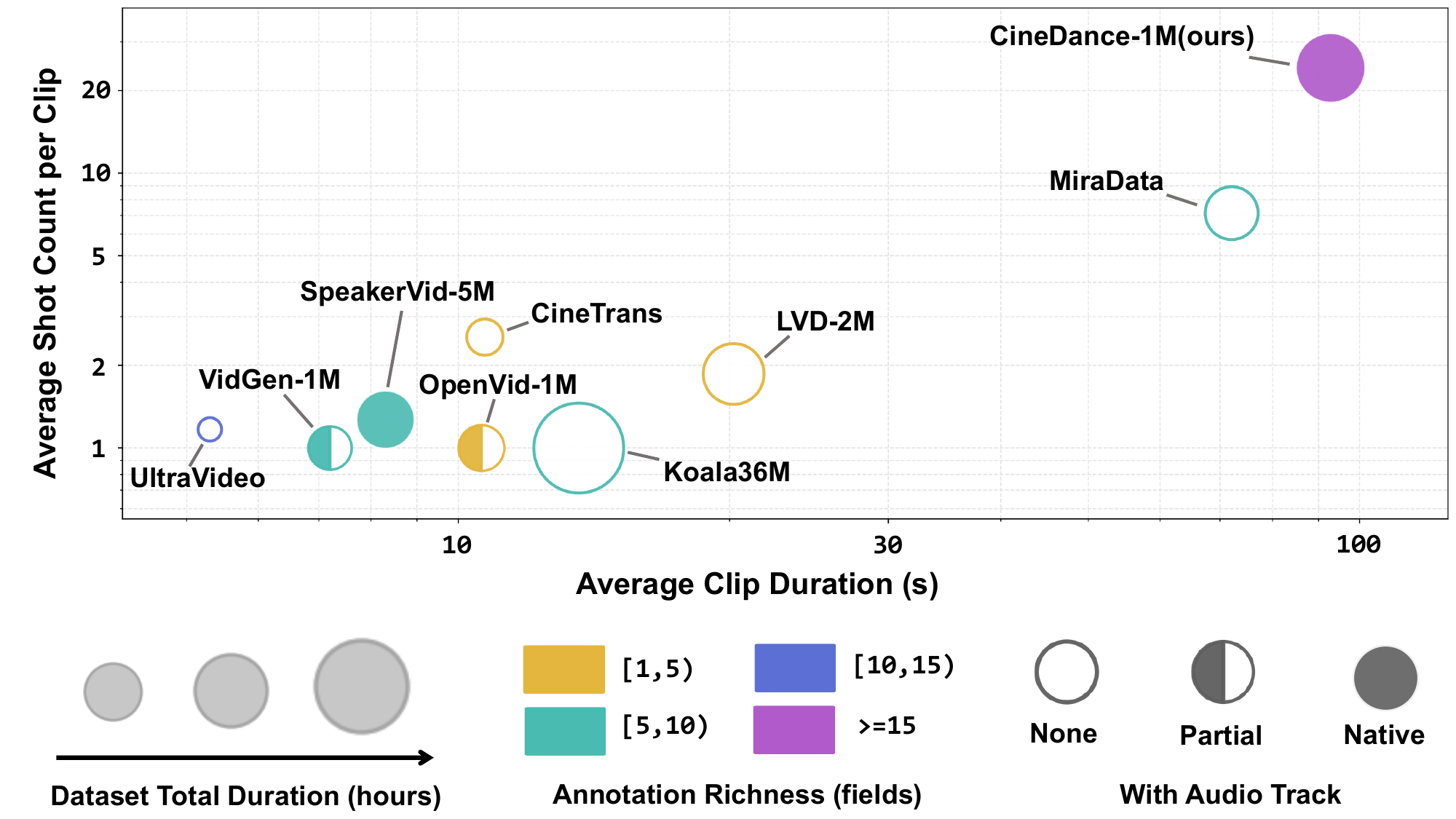}
        \caption{Dataset comparison in scale, average duration, shot count, annotation richness, and audio availability.}
        \label{fig:dataset_gap}
    \end{subfigure}

    \caption{
    (a) Representative T2V and joint audio-video foundation models show clear quality degradation when extended from 5s to 30s.
    (b) Existing datasets remain limited in average clip duration, narrative complexity, annotation granularity, and audio-track availability.
    }
    \label{fig:intro_motivation}
    \vspace{-10pt}
\end{figure}

These challenges further expose \textbf{\textit{two fundamental gaps}} in the current open research ecosystem.
% This limitation is not merely an engineering issue of generating longer clips, but reflects a deeper mismatch between short-form generative priors and the structured temporal logic of cinematic narratives.
The first is a \textbf{foundation-model gap}.
Although recent T2V and T2AV foundation models achieve impressive short-form generation quality, they still suffer from fragmented capabilities and limited responsiveness to multi-shot prompts when extended to long-form multi-shot audio-video generation.
As shown in Fig.~\ref{fig:intro_motivation}(\subref{fig:radar}), representative open video-only models and native audio-video models exhibit substantial degradation when transferring from 5-second generations to 30-second generations.
Except for motion smoothness, which remains relatively high partly because the generated videos often become less dynamic, most other dimensions decrease substantially, including visual quality, audio quality, prompt alignment, and audio-video synchronization, consistent with the observation in Fig.~\ref{fig:intro_failure_cases}.
This suggests that current foundation models still struggle to maintain holistic audio-video quality over extended cinematic sequences.
The second is a \textbf{data-and-benchmark gap}.
\textit{For training}, existing video datasets are typically limited by several factors: short average clip duration, weak narrative complexity, coarse annotation granularity, or the lack of native audio tracks.
Recent efforts such as LVD-2M~\cite{lvd-2m} and MiraData~\cite{miradata} mark important progress toward longer-duration and multi-shot video data, yet they still lack native audio tracks and shot-level dual-modal dense annotations.
Moreover, their narrative complexity remains somewhat limited due to their parsing and curation strategies.
As summarized in Fig.~\ref{fig:intro_motivation}(\subref{fig:dataset_gap}), existing datasets therefore provide limited supervision for long-form multi-shot T2AV generation.
\textit{For evaluation}, a similar limitation exists.
Most previous benchmarks focus on single-shot video-only evaluation, as shown in Tab.~\ref{tab:intro_benchmark_gap}.
Although recent multi-shot benchmarks such as MSVBench~\cite{msvbench} begin to evaluate cross-shot narrative coherence, they remain video-centric, relatively small in scale, and do not provide unified joint audio-video evaluation.
Together, these data and benchmark limitations motivate CineDance-1M, a large-scale structured dataset for multi-shot T2AV generation, and CineBench, a unified benchmark for evaluating long-form multi-shot audio-video narratives.

\begin{table*}[t]
\centering
\caption{
Comparison of representative generation benchmarks (one per category from Sec.~\ref{sec24}).
\BenchYes/\BenchPart/\BenchNo ~denote explicit/partial/absent coverage.
% Size denotes the number of prompts or story scripts.
% Video: video-quality evaluation, including appearance, motion, and temporal quality;
% Audio: audio-quality evaluation;
AV-Sync: audio-video synchronization or cross-modal alignment;
MultiShot: multi-shot structure;
Narr.Cont.: cross-shot narrative continuity;
Struct.Prompt: structured shot-level prompt.
}
\label{tab:intro_benchmark_gap}
\begin{tabular}{lcccccccc}
\toprule
Benchmark 
& Year
& Size
& Video
& Audio
& AV-Sync
& MultiShot
& Narr.Cont.
& Struct.Prompt \\
\midrule
VBench-series~\cite{vbench,vbench++}
& 2024
& 1600
& \BenchYes
& \BenchNo
& \BenchNo
& \BenchPart
& \BenchNo
& \BenchNo \\

FETV~\cite{fetv}
& 2023
& 619
& \BenchPart
& \BenchNo
& \BenchNo
& \BenchNo
& \BenchNo
& \BenchNo \\

T2V-CompBench~\cite{t2v-compbench}
& 2025
& 1400
& \BenchPart
& \BenchNo
& \BenchNo
& \BenchNo
& \BenchNo
& \BenchPart \\

AVGen-Bench~\cite{avgen-bench}
& 2026
& 235
& \BenchYes
& \BenchYes
& \BenchYes
& \BenchNo
& \BenchNo
& \BenchNo \\

MSVBench~\cite{msvbench}
& 2026
& 20
& \BenchPart
& \BenchNo
& \BenchNo
& \BenchYes
& \BenchYes
& \BenchYes \\

\midrule
\textbf{CineBench (ours)}
& 2026
& 1000
& \BenchYes
& \BenchYes
& \BenchYes
& \BenchYes
& \BenchYes
& \BenchYes \\
\bottomrule
\end{tabular}
\end{table*}

To bridge this gap, we introduce \textbf{CineDance-1M}, the first large-scale T2AV dataset designed to catalyze the paradigm shift towards multi-shot long-form generation.
Derived from premium cinematic media, it comprises \textbf{one million} sequences (averaging \AvgDur s and \AvgShots shots) at a minimum 1080p resolution, as shown in Fig.~\ref{fig:teaser}.
Crucially, it provides the first \textbf{natively structured, configurable, and shot-wise annotations for dual modalities}.
The exceptional fidelity of \Ours stems from a rigorous three-stage curation pipeline, which encapsulates our core contributions:
\textbf{1) Diverse Sourcing and Comprehensive Cleansing.} We collect diverse 1080p videos to guarantee pristine generative priors. Following rigorous spatio-temporal cropping and subtitle removal, we comprehensively evaluate visual quality, audio fidelity, and cross-modal alignment to compile a structured metadata dictionary for versatile downstream filtering (Sec.~\ref{sec21}).
\textbf{2) Film-Theory-Inspired Narrative Parsing.}
Moving beyond conventional scene-cutting, we introduce a state-based, bottom-up grouping algorithm. By formalizing cinematic syntax into core parsing and merging rules, we guide Qwen-3.5-27B~\cite{qwen3} to assemble discrete TransNetV2~\cite{transnetv2} shots into coherent, long-form narratives (Sec.~\ref{sec22}).
\textbf{3) Configurable Dual-Modal Annotation.} We propose a hierarchical captioning paradigm using anchor tokens to rigidly bind global subject definitions with granular shot-wise references. We deploy specialized MLLMs (Qwen3.5-35B-A3B~\cite{qwen3} for video and Qwen3-Omni-30B-A3B~\cite{qwen3-omni} for audio) using a task-decomposition strategy, explicitly enabling cross-modal binding while reducing hallucinations (Sec.~\ref{sec23}).
To systematically evaluate complex cinematic synthesis, we introduce \textbf{CineBench}, a comprehensive evaluation suite featuring difficulty-stratified prompt tiers and novel annotation-grounded metrics designed to accurately measure cross-shot narrative continuity and joint audio-video alignment.
To verify the effectiveness of \Ours, we extend the LTX-2.3 model~\cite{ltx-2} to \textbf{\OurModel}, establishing a strong baseline that exhibits robust capabilities in multi-shot, long-form joint audio-video generation.
In summary, our core contributions are fourfold: \\
\hspace*{1em} \textbf{\textit{1)}} Addressing the critical scarcity of narrative-driven data, we introduce \Ours, the first large-scale, 1080p dataset dedicated to multi-shot long-form joint audio-video generation, effectively bridging the gap between single-shot clips and complex cinematic synthesis. \\
\hspace*{1em} \textbf{\textit{2)}} Empowered by a highly automated, structure-aware processing pipeline, our approach integrates film-theory-inspired narrative parsing with hierarchical dual-modal captioning to guarantee premium caption quality, unprecedented semantic density, and precise cross-modal alignment. \\
\hspace*{1em} \textbf{\textit{3)}} To systematically quantify generative capabilities in this new paradigm, we establish CineBench, a comprehensive evaluation suite that stratifies benchmark instances by theme and complexity-based difficulty, also employing a robust six-dimensional metric system for holistic assessment. \\
\hspace*{1em} \textbf{\textit{4)}} By fine-tuning the LTX-2.3 backbone on our dataset, we establish \OurModel as a robust baseline that demonstrates significant superiority over existing models in maintaining spatio-temporal consistency, identity preservation, and precise cross-modal synchronization.

\section{Related Works}\label{relatedworks}

% \noindent\textbf{Open-Source Datasets.}
\subsection{Open-Source Dataset}
The evolution of generative models has been fundamentally propelled by
the availability of large-scale datasets. Early pioneering collections
such as HowTo100M~\cite{howto100m}, WebVid-10M~\cite{webvid-10m}, InternVid~\cite{internvid}, and Panda-70M~\cite{panda70m} prioritized
massive scale and highly diverse web sourcing, laying a crucial
foundation that significantly accelerated the development of early
models. Recent datasets including OpenHumanVid~\cite{openhumanvid}, Koala-36M~\cite{koala}, OpenVid~\cite{openvid},
VideoUFO~\cite{videoufo}, and UltraVideo~\cite{ultravideo} utilize advanced structured captioning and
higher spatial resolutions to enhance visual fidelity. Concurrently,
collections like MiraData~\cite{miradata} and LVD-2M~\cite{lvd-2m} explore long-form video generation.
However, constrained by their data sourcing and basic shot-merging
strategies, their videos remain relatively short and predominantly
single-shot. In the audio-visual domain, large-scale datasets such as
VGGSound~\cite{vggsound} and SpeakerVid-5M~\cite{speakervid-5m} provide valuable cross-modal pairs for
general alignment and human-centric generation, with another line of
works focusing on highly specialized talking-head scenarios like
VoxCeleb~\cite{voxceleb} and LRS3~\cite{lrs3}. Despite their respective contributions, the
aforementioned open-source datasets suffer to varying degrees from
critical limitations, including brief durations, single-shot dynamics,
the absence of the acoustic modality, and a lack of configurable
structured annotations.

\subsection{Foundation Models and Joint Audio-Video Generation}

\noindent\textbf{Video foundation models.}
Recent video foundation models have substantially advanced open-domain text-to-video (T2V) and image-to-video (I2V) generation. 
Representative systems such as HunyuanVideo~\cite{hunyuanvideo,hunyuanvideo-1.5}, Wan~\cite{wan}, SkyReels~\cite{skyreels-v2}, CogVideoX~\cite{cogvideox} and Open-Sora~\cite{opensora} improve visual fidelity, motion quality, and prompt following through stronger diffusion backbones, large-scale training data, and post-training strategies. 
These models provide powerful short-form visual priors, but most remain video-only and are optimized for short clips rather than structured multi-shot narratives. 
When directly extended to long-form cinematic generation, they often suffer from temporal degradation, weakened motion dynamics, and cross-shot identity or scene drift, indicating that strong short-clip synthesis does not automatically translate into long-form narrative control.

\noindent\textbf{Joint audio-video generation.}
Recent works further explore audio-video generative modeling beyond silent videos. 
MMAudio~\cite{mmaudio} synthesizes synchronized audio from video and optional text conditions via multimodal joint training. 
JavisDiT introduces a joint audio-video diffusion transformer with hierarchical spatio-temporal synchronization priors~\cite{javisdit}, while Ovi~\cite{ovi} models audio and video as a unified generative process through twin DiT backbones and blockwise cross-modal fusion. 
Other systems such as Harmony~\cite{harmony}, UniVerse~\cite{universe}, and LTX-2.3~\cite{ltx-2} further study synchronized speech, sound effects, and video generation in increasingly unified frameworks. 
Despite this progress, existing works are still mostly evaluated on short clips or local audio-video alignment, and rarely address long-form multi-shot cinematic narratives that require recurring speaker binding, voice timbre consistency, ambient sound continuity, and shot-level structured control.

% \noindent\textbf{Multi-Shot Long-Form Video Generation.}
\subsection{Multi-Shot Long-Form Video Generation}
\label{sec:rw3}
When extending single-shot foundation models to multi-shot scenarios,
prevalent strategies involve employing masked attention mechanisms~\cite{captioncinema,cinetrans,mask2dit,shotadapter} to
explicitly isolate and distinguish distinct shots, alongside structural
modifications like Rotary Position Embeddings (RoPE)~\cite{rope} to enforce
temporal boundaries~\cite{multishotmaster,shotverse}. Transitioning to long-form generation, shot-by-shot
paradigms~\cite{storymem,filmweaver} have been extensively explored to synthesize individual
segments within a single forward pass. Aimed at efficiently modeling
global consistency under strict computational budgets, sparse attention
patterns are actively integrated~\cite{holocine,moga}, while emerging autoregressive
frameworks\cite{filmweaver,helios,memflow} have concurrently garnered widespread attention for their
ability to generate continuous sequences. Furthermore, to explicitly
correlate distinct shots and maintain long-term visual consistency and
semantic coherence, keyframe-conditioned generation~\cite{stage,captioncinema,cineverse,vgot,movieagent,dreamfactory,moviedreamer,omniweaving,onestory} is widely utilized,
frequently coupled with Vision-Language Models (VLMs) or Multimodal
Large Language Models (MLLMs) to orchestrate narrative progression.
More generally, memory-driven approaches~\cite{memflow,storymem,filmweaver,captioncinema,shotstream} preserve critical feature
representations from preceding shots, leveraging them as conditional
anchors to seamlessly guide subsequent generation steps.
Despite these advances, existing approaches still largely rely on external structural priors to impose multi-shot organization at inference time.
% As a result, how to internalize shot transitions, long-range narrative continuity, and joint audio-video consistency into a native end-to-end generative model remains underexplored.

\begin{figure*}[t]
    \centering
    \includegraphics[width=\linewidth]{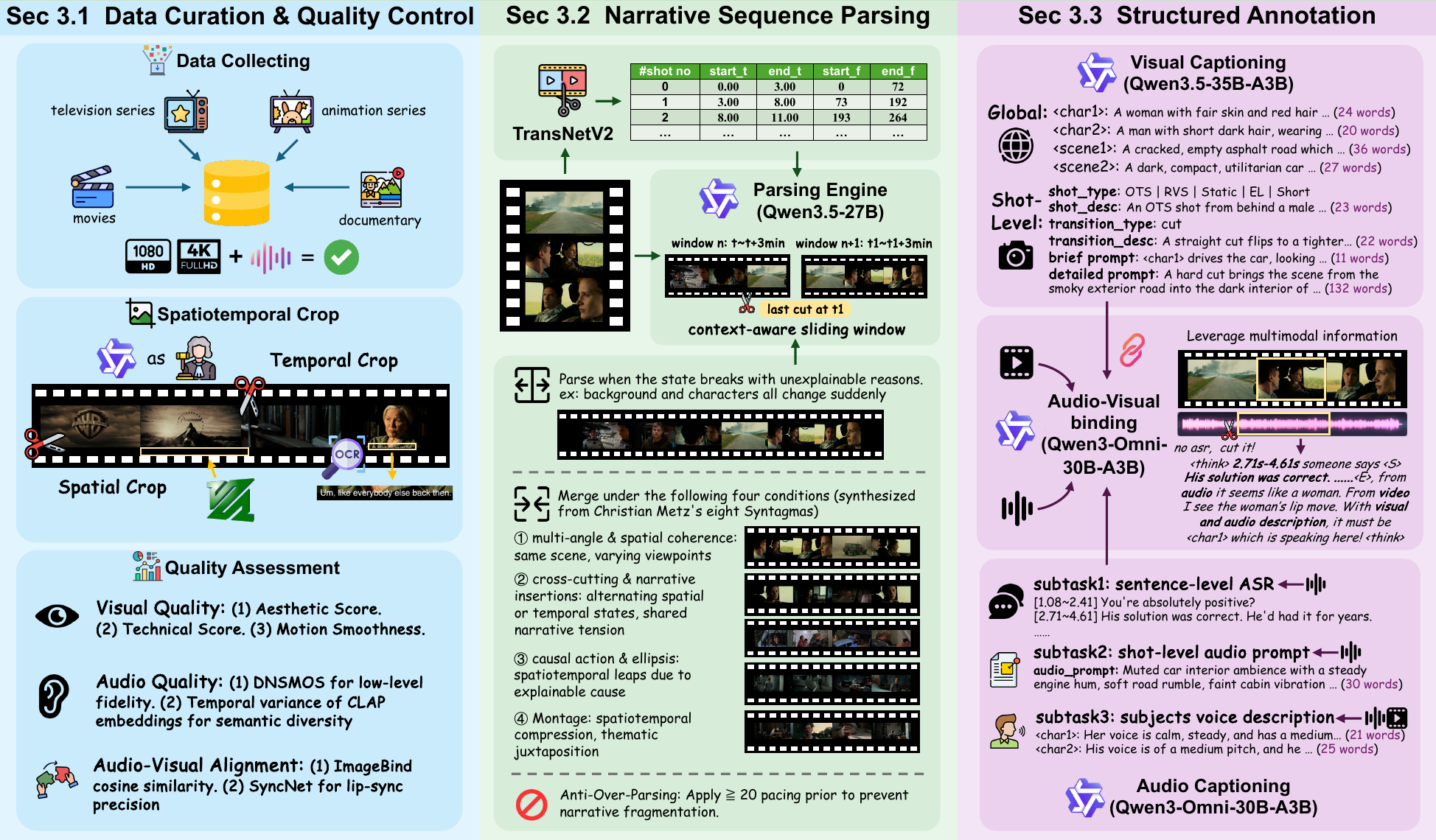}
    \caption{\Ours curation pipeline consists of three main stages: \textbf{\textit{1)}} Data Preparation and Quality Assessment (Sec.~\ref{sec21}), which encompasses data sourcing, spatiotemporal cropping, and quality assessment; \textbf{\textit{2)}} Multi-shot Narrative Parsing (Sec.~\ref{sec22}); and \textbf{\textit{3)}} Hierarchical and Configurable Dual-modal Annotation (Sec.~\ref{sec23}). }
    \label{fig:pipeline}
\end{figure*}

\subsection{Benchmarks for Video and Audio-Video Generation}
\label{sec24}
\textbf{\textit{1)}} \textit{General video generation evaluation.}
Benchmarks such as VBench/VBench++~\cite{vbench,vbench++,vbench2} evaluate video generation through hierarchical dimensions including visual quality, temporal consistency, motion, subject consistency, and prompt alignment. 
EvalCrafter~\cite{evalcrafter}, VideoScore~\cite{videoscore,videoscore2}, FETV~\cite{fetv}, and AIGCBench~\cite{aigcbench} further introduce diverse prompts, human-aligned scoring, fine-grained prompt taxonomies, or controllability-oriented metrics. 
However, they mainly target short-form visual generation rather than long-form multi-shot audio-video narratives.
\textbf{\textit{2)}} \textit{Compositional and prompt-centric evaluation.}
T2V-CompBench~\cite{t2v-compbench} evaluates compositional prompt following, including attribute binding, spatial relations, actions, interactions, and numeracy. 
Other benchmarks study specific capabilities such as text rendering, temporal metamorphosis, motion perception, video dynamics, or real-user prompt distributions~\cite{t2vtextbench,chronomagic-bench,vmbench,devil,vidprom}. 
These diagnostics are valuable, but mostly remain short-form or single-shot and lack cross-shot narrative evaluation.
\textbf{\textit{3)}} \textit{T2AV evaluation.}
TAVGBench~\cite{tavgbench} and AVGen-Bench~\cite{avgen-bench} evaluate text-to-audio-video generation with emphasis on cross-modal alignment, unimodal quality, and semantic controllability. 
MTAVG-Bench~\cite{mtavg-bench}, VidAudio-Bench~\cite{vidaudio-bench}, and VABench~\cite{vabench} further cover multi-speaker dialogue, video-to-audio, video-text-to-audio, or synchronous-audio video generation. 
They advance audio-video evaluation, but mainly focus on local AV alignment or speech coherence rather than long-form cinematic narratives.
\textbf{\textit{4)}} \textit{Multi-shot and narrative evaluation.}
MSVBench~\cite{msvbench} evaluates multi-shot generation with hierarchical scripts, reference images, and cross-shot metrics; EntityBench~\cite{entitybench} studies long-range consistency of characters, objects, and locations; and MuSS~\cite{muss} targets multi-shot subject-to-video narrative evaluation. 
These benchmarks approach narrative generation, but remain largely video-centric or subject-to-video oriented. 
% CineBench targets the missing intersection: structured long-form multi-shot audio-video generation with joint evaluation of video quality, audio quality, AV synchronization, prompt alignment, and cross-shot narrative continuity.

\section{Curating CineDance-1M for Cinematic Audio-Video Generation}
\label{sec2}
While recent datasets predominantly focus on the massive scale of discrete short clips, \Ours prioritizes high-fidelity, narrative-driven cinematic sequences.
As intuitively outlined in Fig.~\ref{fig:pipeline}, our rigorous data curation pipeline consists of three core stages: \textbf{\textit{1)}} Data Preparation and Quality Assessment (Sec.~\ref{sec21}), \textbf{\textit{2)}} Bottom-Up Narrative Sequence Parsing via State-Based Shot Grouping (Sec.~\ref{sec22}), and \textbf{\textit{3)}} Configurable Structured Dual-Modal Annotation (Sec.~\ref{sec23}).
To rigorously validate the pipeline's efficacy and optimality, and to ensure the exceptional quality of our dataset, we conduct comprehensive benchmarking across various implementations with in-depth discussions. 
% Furthermore, we execute thorough human evaluations on the processed outcomes of all three stages, with all detailed evidence provided in the supplementary material (Appendices \ref{app:pipeline-data-format}, \ref{app:pipeline-narrative-parsing}, and \ref{app:pipeline-annotation}).

\begin{table}[t]
\centering
\caption{Manual visual artifact audit on 500 random short clips per corpus. 
A clip is counted as non-compliant if it contains at least one residual artifact targeted by our filtering stage.}
\label{tab:visual_artifact_audit}
% \begingroup
% \setlength{\tabcolsep}{3.5pt}
% \renewcommand{\arraystretch}{0.95}
% \small
\begin{tabular}{lccc}
\toprule
Corpus & \#Clips & \#Artifacts & Rate \\
\midrule
Koala-36M~\cite{koala} & 500 & 187 & 37.4\% \\
\textbf{CineDance-1M (ours)} & 500 & 14 & 2.8\% \\
\bottomrule
\end{tabular}
% \endgroup
\end{table}

\begin{table*}[t]
\centering
\caption{
Curation funnel of CineDance-1M. 
Duration denotes the retained temporal coverage after each stage.
Quality assessment computes and stores metadata only, without filtering sequences by default.
}
\label{tab:curation_funnel}
\setlength{\tabcolsep}{2.7pt}
\renewcommand{\arraystretch}{1.0}
% \small
\begin{tabular}{@{}llrrp{0.47\textwidth}@{}}
\toprule
Stage & Unit & Count & Duration & Operation \\
\midrule
Raw collection (Sec.~\ref{sec21})
& Videos & \RawFiles & 32.8K hr 
& Collect 1080p source videos with provenance records \\

Spatiotemporal pre-filter (Sec.~\ref{sec21})
& Videos & 44,579 & 32.5K hr 
& Remove subtitles, borders, title cards, openings, and credits \\

Shot detection (Sec.~\ref{sec22})
& Shots & 25,899,474 & 32.5K hr 
& Detect atomic shot boundaries using TransNetV2 \\

Narrative parsing (Sec.~\ref{sec22})
& Seq. & 1,201,912 & 32.5K hr 
& Group shots into state-consistent narrative sequences \\

Sequence pruning (Sec.~\ref{sec21})
& Seq. & 1,079,382 & 28.6K hr 
& Remove sequences that are single-shot or shorter than 10s \\

Post-verification (Sec.~\ref{sec21})
& Seq. 
& \multirow{2}{*}{1,021,657}
& \multirow{2}{*}{26.3K hr}
& Reject temporally invalid or artifact-contaminated sequences \\

Quality assessment (Sec.~\ref{sec21})
& Seq. 
& 
& 
& Metric-based quality assessment without actual pruning \\
\bottomrule
\end{tabular}
\end{table*}

\subsection{Data Preparation and Quality Assessment}
\label{sec21}

\noindent\textbf{Source data collection.}
The raw video corpus of \Ours is constructed from two principal sources: \textbf{\textit{1)}} we extract source data from widely adopted public datasets, including MiraData~\cite{miradata}, LVD-2M~\cite{lvd-2m}, and Koala36M~\cite{koala}, applying stringent duration constraints and multi-shot filtering; and \textbf{\textit{2)}} we augment this corpus by adhering to the established data collection pipelines of SkyReels-V2~\cite{skyreels-v2} and OpenHumanVid~\cite{openhumanvid}. Subsequently, we conduct a thorough manual curation process to exclude low-quality samples and potentially sensitive content. The finalized source set consists of \textbf{\RawFiles~raw videos}, each with a minimum spatial resolution of 1080p and a cumulative duration exceeding \textbf{\VideoHours~hours}. The inherent narrative coherence, sophisticated visual composition, and high-fidelity acoustic characteristics of these videos render this corpus particularly well-suited for advancing research on joint audio--video generative models.

\noindent\textbf{Coarse-to-fine spatiotemporal filtering.}
To remove black borders and overlaid text, we apply a coarse-to-fine spatiotemporal filtering and cropping pipeline to the raw videos.
The first two steps operate at the raw-video level in a coarse manner and serve as preprocessing for the narrative parsing stage in Sec.~\ref{sec22}: they reduce the interference of letterboxing, subtitles, title cards, openings, and end credits on shot boundary detection and semantic parsing.
After raw-video parsing (Sec.~\ref{sec22}), we further perform fine clip-level verification to further improve the quality of the final clips.
The pipeline is divided into three steps:  \\
\hspace*{1em} \noindent \textit{\textbf{1) Coarse Spatial Cropping.}} To remove global subtitles and letterboxing, we sample segments at the 25\%, 50\%, and 75\% milestones at 2 FPS. Using \texttt{EasyOCR}~\cite{easyocr} for text detection and \texttt{FFmpeg} for black border detection, we compute the optimal bounding boxes and perform a global spatial crop.\\
\hspace*{1em} \noindent \textit{\textbf{2) MLLM-Guided Temporal Truncation.}} We feed the initial and final segments of duration $t = \max(5\text{ min}, 0.1L)$, where $L$ denotes the total video length, to MLLMs to accurately detect and remove non-narrative introductory and concluding content. If the model judges an entire queried segment as introductory or concluding, we iteratively move inward and query the next segment of the same duration until a narrative boundary is identified.\\
\hspace*{1em} \noindent \textit{\textbf{3) Fine Clip-Level Verification.}} To prevent under- or over-cropping caused by dynamic aspect ratios (e.g., IMAX transitions) and avoid missing intermittent text, we additionally perform clip-level OCR and black border detection after the raw video parsing detailed in Sec.~\ref{sec22}. 
% Furthermore, clips exceeding a text density threshold, measured by the average frame-level text-area ratio, are discarded.
We discard clips whose average frame-level text-area ratio exceeds a predefined threshold.

\noindent\textbf{Metric-based quality assessment.}
To systematically ensure the dataset's fidelity and enable versatile downstream filtering, we evaluate each final video clip across three core dimensions following raw video parsing (detailed in Sec.~\ref{sec22}):
\textit{\textbf{1) Video Quality.}} We systematically evaluate \textit{Aesthetic Quality} and \textit{Technical Score} (DOVER~\cite{dover}), along with \textit{Motion Smoothness} (AMT~\cite{amt}) at both the shot and video levels.
\textit{\textbf{2) Audio Quality.}} Apart from low-level signal fidelity (DNSMOS~\cite{dnsmos}), we assess acoustic richness by measuring the temporal variance of CLAP~\cite{clap} embeddings, both evaluated at the video level.
\textit{\textbf{3) Audio-Video Alignment.}} We quantify cross-modal consistency via ImageBind~\cite{imagebind} for global audio-video alignment, and employ SyncNet~\cite{syncnet} to measure lip-synchronization. \\
Instead of applying hard pruning with fixed thresholds, we store all quality scores as metadata, enabling users to flexibly construct task-specific subsets and control the quantity--quality trade-off.  \\
\noindent\textbf{Visual artifact audit.}
To verify that the coarse-to-fine filtering stage removes the visual artifacts it targets, we conduct a manual audit against Koala-36M, a recent large-scale and high-quality T2V dataset. 
We randomly sample 500 clips from CineDance-1M and Koala-36M, respectively, and ask three trained annotators to independently inspect residual artifacts, including burnt subtitles or logos, letterboxing, watermarks, network overlays, title-card or end-credit frames, screen recordings, transition effects, still-frame holds, and near-uniform fill frames. 
A clip is counted as non-compliant with our artifact-removal criteria if any of these artifacts appears; annotator disagreements are resolved by joint review. 
Although CineDance-1M clips are substantially longer on average than Koala-36M clips, CineDance-1M reduces the non-compliance rate from 37.4\% on Koala-36M to 2.8\%, a 13.4$\times$ reduction, as shown in Tab.~\ref{tab:visual_artifact_audit}. 
The remaining failures are mainly intermittent watermark frames, which appear only briefly and are difficult to remove without overly aggressive cropping.

% preamble:
% \usepackage{booktabs}
% \usepackage{multirow}
% \usepackage{array}

% \begin{table*}[t]
% \centering
% \caption{Curation funnel of CineDance-1M. Duration denotes retained temporal coverage after each stage; quality assessment stores metadata and does not remove sequences by default.}
% \label{tab:curation_funnel}
% \begingroup
% \setlength{\tabcolsep}{4pt}
% \renewcommand{\arraystretch}{0.98}
% \small
% \begin{tabularx}{0.78\textwidth}{llrrX}
% \toprule
% Stage & Unit & Count & Dur. & Output / criterion \\
% \midrule
% Raw collection & Videos & 42,876 & 31.5K hr & 1080P source videos with provenance records \\
% ST pre-filter & Videos & 42,876 & -- & Remove subtitles, borders, title cards, openings, and credits \\
% Shot detection & Shots & -- & -- & Detect atomic shot boundaries with TransNetV2 \\
% Narrative parsing & Seq. & -- & -- & Group shots into state-consistent narrative sequences \\
% Seq. pruning & Seq. & -- & -- & Remove single-shot sequences and sequences shorter than 10s \\
% Post-verification & Seq. & -- & -- & Reject temporally invalid or artifact-contaminated sequences \\
% Quality metadata & Seq. & same & same & Store visual, audio, and AV quality scores \\
% \bottomrule
% \end{tabularx}
% \endgroup
% \end{table*}

\begin{table*}[t]
\centering
\caption{
Narrative parsing ablation on a 12-film, 25.2-hour human reference set.
``Align.'' measures boundary alignment with TransNetV2 shot cuts.
\textbf{Bold} indicates the best machine result, $^\star$ denotes the selected implementation, and $^\dagger$ marks the empirical basis of the \textbf{\textit{20s anti-fragmentation prior}}.
}
\label{tab:narrative_parsing_ablation}
\begingroup
\setlength{\tabcolsep}{3.2pt}
\renewcommand{\arraystretch}{0.98}
\small
\begin{tabular}{llrrrrrr}
\toprule
Source / Backbone & Strategy & \#Seq. & Mean (s) & Min (s) & $<20$s (\%) $\downarrow$ & Align. (\%) $\uparrow$ & F1 (\%) $\uparrow$ \\
\midrule
Human GT 
& Reconciled annotation 
& 842 & 108.3 & \textbf{\textit{18.4}}$^\dagger$ & 0.6 & -- & -- \\

\midrule
\multirow{2}{*}{Qwen3.5-35B-A3B} 
& Direct timestamp parsing 
& 1,748 & 51.6 & 2.1 & 36.2 & 53.2 & 51.7 \\
& Bottom-up shot grouping 
& 1,347 & 67.6 & 5.8 & 22.3 & \textbf{100.0} & 66.9 \\

\midrule
\multirow{2}{*}{Qwen3.5-122B-A10B} 
& Direct timestamp parsing 
& 1,342 & 67.4 & 3.7 & 24.3 & 64.4 & 62.7 \\
& Bottom-up shot grouping 
& 1,082 & 84.1 & 9.5 & 11.7 & \textbf{100.0} & 77.4 \\

\midrule
\multirow{2}{*}{Qwen3.5-27B} 
& Direct timestamp parsing 
& 1,078 & 84.2 & 6.4 & 16.2 & 70.6 & 73.4 \\
& \textbf{Bottom-up shot grouping$^\star$}
& 873 & 104.4 & 17.2 & \textbf{3.1} & \textbf{100.0} & \textbf{88.4} \\
\bottomrule
\end{tabular}
\endgroup
\end{table*}

\subsection{Bottom-Up Narrative Sequence Parsing via State-Based Shot Grouping}
\label{sec22}

\noindent\textbf{Narrative sequences definition.}
Existing long-video datasets~\cite{miradata,moviebench,cinetrans,lvd-2m} typically rely on low-level visual cues (e.g., pixel similarity) for video segmentation. However, this rigid strategy disrupts inherent narrative continuity, as physical scene transitions do not necessarily signal narrative breaks (e.g., a continuous conversation that moves from indoors to outdoors). To preserve cinematic cohesion, we formally define a \textbf{\textit{Narrative Sequence}} as a continuous flow of diegetic time and causality that aligns with real-world chronological progression, regardless of spatial shifts. Within this unit, all character and environmental state changes are strictly driven by continuous logical events and clear causal explanations (exceptions are detailed in the following sections).

\noindent\textbf{State-based parsing rule.} Based on our previous definition, we conceptually model the cinematic flow as a transition of semantic states, $S = (\mathcal{T}, \mathcal{P}, \mathcal{C}, \mathcal{E})$, representing time, place, characters, and events. We establish the \textbf{\textit{first-principle parsing rule}}: a narrative sequence terminates when an \textit{unexplainable state break} occurs (\textit{e.g.}, abrupt shifts in character identity, total replacement of the character ensemble, or complete change of the environmental layout).

\noindent\textbf{Cinematic-theory-guided merging rules.}
However, while a state break is a necessary condition for a narrative boundary, it is not a sufficient one.
As illustrated in Fig.~\ref{fig:teaser}, cross-cutting shots and montage techniques~\cite{montage} are frequently employed in cinematic storytelling.
To address this, inspired by Christian Metz's \textit{Grande Syntagmatique}~\cite{grande}, we adapt and filter classical film syntax into \textbf{\textit{four merging rules}} to instruct the MLLM not to cut during these cinematic exceptions: \\
\hspace*{1em} \noindent  \textit{\textbf{1) Multi-Angle and Spatial Coherence:}} Shots exhibiting $\Delta\mathcal{P} = \text{camera-angle-only}$ while maintaining $\Delta\mathcal{T} \approx 0$ and a unified ongoing event ($\mathcal{E}_{continuous}$) are merged, representing continuous action within the same physical environment explored from different viewpoints.\\
\hspace*{1em} \noindent \textit{\textbf{2) Cross-Cutting and Narrative Insertions:}} Sequences rapidly alternating between two distinct spatial states ($\mathcal{P}_A \neq \mathcal{P}_B$) or temporal states ($\mathcal{T}_A \neq \mathcal{T}_B$) are merged if they are bound by a unified causal tension ($\mathcal{E}_{shared}$). This encapsulates simultaneous alternating (e.g., a phone call) and sandwich insertions (e.g., a brief flashback).\\
\hspace*{1em} \noindent \textit{\textbf{3) Causal Action and Ellipsis:}} Shots exhibiting significant spatial leaps ($\Delta\mathcal{P} \neq 0$) and temporal gaps ($\Delta\mathcal{T} > 0$) are merged if the event $\mathcal{E}_{i+1}$ is a direct, explainable causal consequence of $\mathcal{E}_i$ (e.g., a gun fired indoors causing a window to shatter outdoors).\\
\hspace*{1em} \noindent \textit{\textbf{4) Montage:}} Sequences exhibiting disjointed shifts in time and space without explicit micro-causal links are merged if they are unified by a macro-level thematic or emotional arc ($\mathcal{E}_{theme}$), such as a montage of a city waking up or a training sequence.

\noindent\textbf{Empirical anti-fragmentation rule.}
Classical continuity editing emphasizes preserving coherent spatial and temporal relations across shots~\cite{film_art}.
Motivated by this principle, we introduce an explicit \textit{anti-fragmentation rule} to prevent MLLMs from splitting a cohesive narrative segment into meaningless short clips.
As shown in Tab.~\ref{tab:narrative_parsing_ablation}, our reconciled human parsing results give a minimum sequence duration of 18.4 seconds.
We therefore round this empirical lower-tail value to \textbf{20} seconds and use it as a soft minimum-duration prior during MLLM parsing.
When a candidate boundary would produce a segment shorter than this threshold, the parser is instructed to keep expanding the temporal window unless a clear character, scene, or event-state break is observed.

\noindent \textbf{Bottom-up sliding inference.} We first formulate our parsing, merging, and anti-fragmentation rules as structured prompts. Based on comprehensive benchmarks of the Qwen3.5 family~\cite{qwen3}, we then adopt the Qwen3.5-27B model as our core parsing engine, equipped with two core mechanisms:\\
\hspace*{1em} \noindent \textit{\textbf{1) Bottom-up shot indexing.}} Directly prompting an MLLM to localize narrative boundaries across long videos causes severe temporal hallucination. Therefore, we establish shots (extracted via \texttt{TransNetV2}~\cite{transnetv2}) as the atomic units of our dataset. Given the start and end timestamps of each shot, the MLLM is then tasked solely with outputting discrete shot indices as parsing boundaries based on our prompts, significantly reducing timestamp hallucinations and improving parsing accuracy.\\
% \hspace*{1em} \noindent \textit{\textbf{2) Context-aware sliding window inference.}} Given the massive duration of raw videos and MLLM context window constraints, processing entire movies simultaneously is intractable. To resolve this, we implement a context-aware sliding window strategy. Specifically, we sequentially feed 3-minute windows of the video (starting from the narrative onset determined by temporal truncation in Sec.~\ref{sec21}) into the MLLM, and the last detected boundary of the current window serves as the starting point for the next window. If the MLLM outputs no boundaries, indicating the current window is a continuous narrative sequence, we iteratively expand the window size in 3-minute increments until a valid boundary is identified. This dynamic windowing strategy ensures global narrative coherence while maintaining high computational efficiency.
\hspace*{1em}\noindent\textit{\textbf{2) Context-aware sliding window inference.}} 
Given the massive duration of raw videos and MLLM context window constraints, processing entire movies simultaneously is intractable.
To resolve this, we implement a context-aware sliding window strategy, as summarized in Alg.~\ref{alg:narrative_parsing}.
Specifically, starting from the narrative onset determined by temporal truncation in Sec.~\ref{sec21}, we sequentially feed approximately $\Delta=3$ minute windows into the MLLM.
Since parsing windows must align with atomic shot boundaries, each window endpoint is selected by $\mathrm{GetWindowEnd}$, which returns the shot boundary whose temporal span is closest to $\Delta$.
The last detected boundary in the current window serves as the starting point of the next window.
If the MLLM outputs no boundary, indicating that the current window is likely a continuous narrative sequence, we call $\mathrm{ExtendWindowEnd}$ to expand the current window by another reference duration $\Delta$, again snapping the endpoint to the nearest shot boundary.
This process is repeated until a valid boundary is identified or the end of the video is reached.
This dynamic windowing strategy preserves global narrative coherence while maintaining high computational efficiency.

\begin{algorithm}[t]
\caption{Bottom-Up Narrative Sequence Parsing}
\label{alg:narrative_parsing}
\KwIn{Raw video $V$; shot detector $D$; parsing engine $M$; 
reference window size $\Delta$; rule set $R=\{R_p,R_m,R_a\}$}
\KwOut{Boundary list $B$ for cropping narrative sequences}

\BlankLine
\SetKwProg{Fn}{Function}{:}{}
\SetKwFunction{GetEnd}{GetWindowEnd}
\SetKwFunction{ExtendEnd}{ExtendWindowEnd}

\Fn{\GetEnd{S, l, $\Delta$}}{
    Find the shot index $r$ such that the boundary-aligned window
    $\{s_l,\ldots,s_r\}$ has duration closest to $\Delta$\;
    \Return{$r$}\;
}

\Fn{\ExtendEnd{S, l, r, $\Delta$}}{
    $r \leftarrow \GetEnd{S, l, $\mathrm{Dur}(S,l,r)+\Delta$}$\;
    \Return{$r$}\;
}

\BlankLine
Detect atomic shots $S=\{s_i\}_{i=1}^{n}$ using $D(V)$\;
Initialize boundary list $B\leftarrow\emptyset$ and window start $l\leftarrow1$\;

\While{$l \leq n$}{
    $r\leftarrow\GetEnd{S, l, $\Delta$}$\;
    $C\leftarrow\emptyset$\;
    
    \While{$C=\emptyset$ \textbf{and} $r<n$}{
        Build window $W=\{s_l,\ldots,s_r\}$ with shot indices, timestamps and sampled frames\;
        Query $M$ with $W$ and $R$ to obtain candidate boundaries $C$\;
        
        \If{$C=\emptyset$}{
            $r\leftarrow\ExtendEnd{S, l, r, $\Delta$}$\;
        }
    }
    
    \If{$C=\emptyset$}{
        $B\leftarrow B\cup\{n+1\}$\;
        \textbf{break}\;
    }
    
    $B\leftarrow B\cup C$\;
    $l\leftarrow\max(C)$\;
}

\Return{$B$}
\end{algorithm}

\subsection{Configurable Structured Dual-Modal Annotation}
\label{sec23}

\noindent\textbf{Configurable anchored design.} Given the diverse prompt formats and input conditions in current multi-shot long-form generation, we introduce an anchor token mechanism to ensure our annotations are highly adaptable for future explorations. Specifically, we define a global character list ($[\langle\texttt{char}_1\rangle, \dots, \langle\texttt{char}_N\rangle]$) and a scene list ($[\langle\texttt{scene}_1\rangle, \dots, \langle\texttt{scene}_M\rangle]$) for the entire narrative sequence. Subsequently, the dual-modal shot-level prompts (detailed in the following sections) explicitly refer back to these anchor tokens. This mechanism significantly enhances prompt configurability while establishing robust connections across global-to-shot and shot-to-shot levels.

\noindent\textbf{Shot-wise structural visual captioning.} For each shot, we extract a five-dimensional set of shot attributes (scale, angle, movement, narrative function, duration category) alongside specific shot transition types~\cite{film_art}. Furthermore, we supplement these categorical labels with a dedicated shot description and a transition description to facilitate advanced training control. Crucially, we define a localized character list and an active scene for each shot, indicating the specific characters present and the environment where the shot takes place. Both the brief and detailed visual prompts then explicitly refer to these anchor tokens. To ensure global consistency and reduce model hallucinations, the global character and scene lists alongside all shot-level visual annotations are generated in a single pass of Qwen3.5-35B-A3B~\cite{qwen3}.

\begin{table}[t]
\centering
\small
\setlength{\tabcolsep}{5pt}
\renewcommand{\arraystretch}{1.12}
\caption{
Audio-only speaker diarization accuracy on our 100-clip cinematic-dialogue benchmark, measured by permutation-invariant segment-level matching averaged across clips.
The results motivate decoupling ASR from global character-level speaker binding.
}
\label{tab:speaker_binding_ablation}
\begin{tabular*}{\linewidth}{@{\extracolsep{\fill}} l l c @{}}
\toprule
Backbone / Family & Implementation & Acc. (\%) \\
\midrule
\multirow{2}{*}{\makecell[l]{Specialized\\diarization}}
& Pyannote-3.1 & 62.7 \\
& DiariZen & 63.1 \\
\addlinespace[2pt]
\midrule
\multirow{2}{*}{\makecell[l]{Closed-source\\MLLM}}
& Gemini-2.5-Pro & 82.8 \\
& Gemini-3.1-Pro & 87.4 \\
\addlinespace[2pt]
\midrule
\multirow{2}{*}{\makecell[l]{Qwen3-Omni\\30B-A3B}}
& Whole-clip prompting & 56.4 \\
& Sliding-window prompting & 83.1 \\
\bottomrule
\end{tabular*}
\end{table}

\begin{table*}[t]
  \centering
  \caption{Comparison of representative video datasets. \Ours pioneers \textit{multi-shot (\AvgShots shots)}, \textit{long-form (\AvgDur s)} T2AV generation with unprecedented structural complexity at 1080p. As the only 1M-scale dataset providing dense shot-level audio-video annotations (averaging > 6,400 words), it uniquely enables granular cross-modal control.}
  \label{tab:dataset-comparison}
  \begin{threeparttable}
    \setlength{\tabcolsep}{4.6pt}
    \renewcommand{\arraystretch}{1.15}
    \small
    \begin{adjustbox}{max width=\textwidth}
      \begin{tabular}{@{}lccccccrrrr@{}}
        \toprule
        \multirow{2}{*}{Dataset} & Visual & \multicolumn{3}{c}{Clip Structure} & \multicolumn{2}{c}{Audio} & Text & \multicolumn{2}{c}{Scale} & Time \\
        \cmidrule(lr){2-2} \cmidrule(lr){3-5} \cmidrule(lr){6-7} \cmidrule(lr){9-10}
         & Res. & Avg. dur. & Avg. shots & Shot caps. & Audio & Audio ann. & Cap. len. & Total dur. & Clips & Year \\
        \midrule
        HowTo100M~\cite{howto100m} & 240p & 3.6s & 1 & None & None & None & 4 & 134.5Khr & 136M & 2019 \\
        HD-VILA-100M~\cite{advancing_high_resolution} & 720p & 13.4s & 1 & None & None & None & 32.5 & 371.5Khr & 103M & 2022 \\
        Koala-36M~\cite{koala} & 720p & 13.6s & 1 & None & None & None & 202.3 & 137Khr & 36M & 2024 \\
        VIDGEN-1M~\cite{vidgen} & 720p & 10.6s & 1 & None & Partial & None & 89.3 & 2.9Khr & 1M & 2024 \\
        MiraData~\cite{miradata} & 720p & 72.1s & 7.15 & None & None & None & 319 & 6.6Khr & 330K & 2024 \\
        LVD-2M~\cite{lvd-2m} & 720p & 20.2s & 1.86 & None & None & None & 88.8 & 14.6Khr & 2.1M & 2024 \\
        OpenHumanVid~\cite{openhumanvid} & 720p & 4.6s & 1 & None & All & None & 99.7 & 12Khr & 16M & 2025 \\
        OpenS2V-5M~\cite{opens2v} & 720p & 5.6s & 1 & None & Partial & None & 312.06 & 5.8Khr & 3.75M & 2025 \\
        UltraVideo~\cite{ultravideo} & 4K/8K & 5.3s & 1.17 & None & No & None & 824.3 & 62hr & 42K & 2025 \\
        OpenVid-1M~\cite{openvid} & 720p & 7.2s & 1 & None & Partial & None & 126.5 & 2.1Khr & 1M & 2025 \\
        CineTrans~\cite{cinetrans} & 720p & 10.7s & 2.53 & 2 & None & None & 250.78 & 752hr & 252K & 2025 \\
        SpeakerVid-5M~\cite{speakervid-5m} & 1080p & 8.3s & 1.27 & None & All & ASR & 20.69 & 11.6Khr & 5.07M & 2025 \\
        \midrule
        \rowcolor{TableAccent}
\textbf{Ours} & \textbf{1080p} & \textbf{\AvgDur s} & \textbf{\AvgShots} & \textbf{22} & \textbf{All} & \textbf{Structured} & \textbf{6496.3} & \textbf{26.3Khr} & \textbf{1M} & \textbf{2026} \\
        \bottomrule
      \end{tabular}
    \end{adjustbox}
  \end{threeparttable}
\end{table*}

\begin{table}[t]
\centering
\small
\setlength{\tabcolsep}{5pt}
\renewcommand{\arraystretch}{1.12}
\caption{
Multimodal ASR segment-to-character binding accuracy on the 100-clip human-labeled benchmark.
Bold indicates our production configuration, and italic indicates the unaffordable closed-source ceiling.
}
\label{tab:av_binding_ablation}
\begin{tabularx}{\linewidth}{@{} X c c @{}}
\toprule
Method & Regime & Acc. (\%) \\
\midrule
\multicolumn{3}{@{}l}{\textit{Open-source omni-MLLMs}} \\
Qwen2.5-Omni-7B-Instruct 
& Whole-clip 
& 58.4 \\
Qwen2.5-Omni-7B-Instruct 
& Windowed 
& 78.6 \\
Qwen3-Omni-30B-A3B-Instruct 
& Whole-clip 
& 67.2 \\
Qwen3-Omni-30B-A3B-Instruct 
& \textbf{Windowed} 
& \textbf{95.4} \\
\addlinespace[2pt]
\midrule
\multicolumn{3}{@{}l}{\textit{Closed-source frontier omni systems}} \\
Gemini-2.5-Flash 
& Whole-clip 
& 64.7 \\
Gemini-2.5-Flash 
& Windowed 
& 88.9 \\
Gemini-2.5-Pro 
& Windowed 
& 92.8 \\
Gemini-3.1-Pro 
& \textit{Windowed} 
& \textit{96.3} \\
\bottomrule
\end{tabularx}
\end{table}

\noindent\textbf{Granular audio diarization and captioning.}
Based on comprehensive benchmarks of various implementations, we decompose the audio annotation process into three manageable sub-tasks via Qwen3-Omni-30B-A3B~\cite{qwen3-omni}: 
\textbf{\textit{1)}} extracting sentence-level Automatic Speech Recognition (ASR) segments, 
\textbf{\textit{2)}} generating shot-level audio prompts, and 
\textbf{\textit{3)}} creating character voice descriptions.
This design is motivated by \textbf{three observations}.
\textbf{\textit{First}}, speech is a relatively special component of audio annotations.
Unlike ambient sound, music, or sound effects, dialogue is often organized in different prompt formats by different joint audio-video generation foundation models~\cite{ovi,ltx-2,mova}.
We therefore store ASR as an independent annotation term, which makes the dataset more flexible for different downstream training and prompting protocols.
\textbf{\textit{Second}}, we do not perform audio-level speaker-to-character binding during ASR.
As shown in Tab.~\ref{tab:speaker_binding_ablation}, existing systems struggle to maintain usable audio-only speaker diarization accuracy (i.e., above 90\%) on long-form cinematic dialogue.
Thus, the ASR stage only extracts spoken text, while global speaker-to-character assignment is handled by the multimodal binding stage described in the next section.
\textbf{\textit{Third}}, even strong open-source omni-modal models such as Qwen3-Omni-30B-A3B can exhibit a high hallucination rate when asked to jointly perform multiple audio-related tasks in a single pass.
By minimizing individual task complexity, this decomposed strategy mitigates hallucinations and improves annotation quality.

\begin{figure*}[t]
    \centering
    \includegraphics[width=\linewidth]{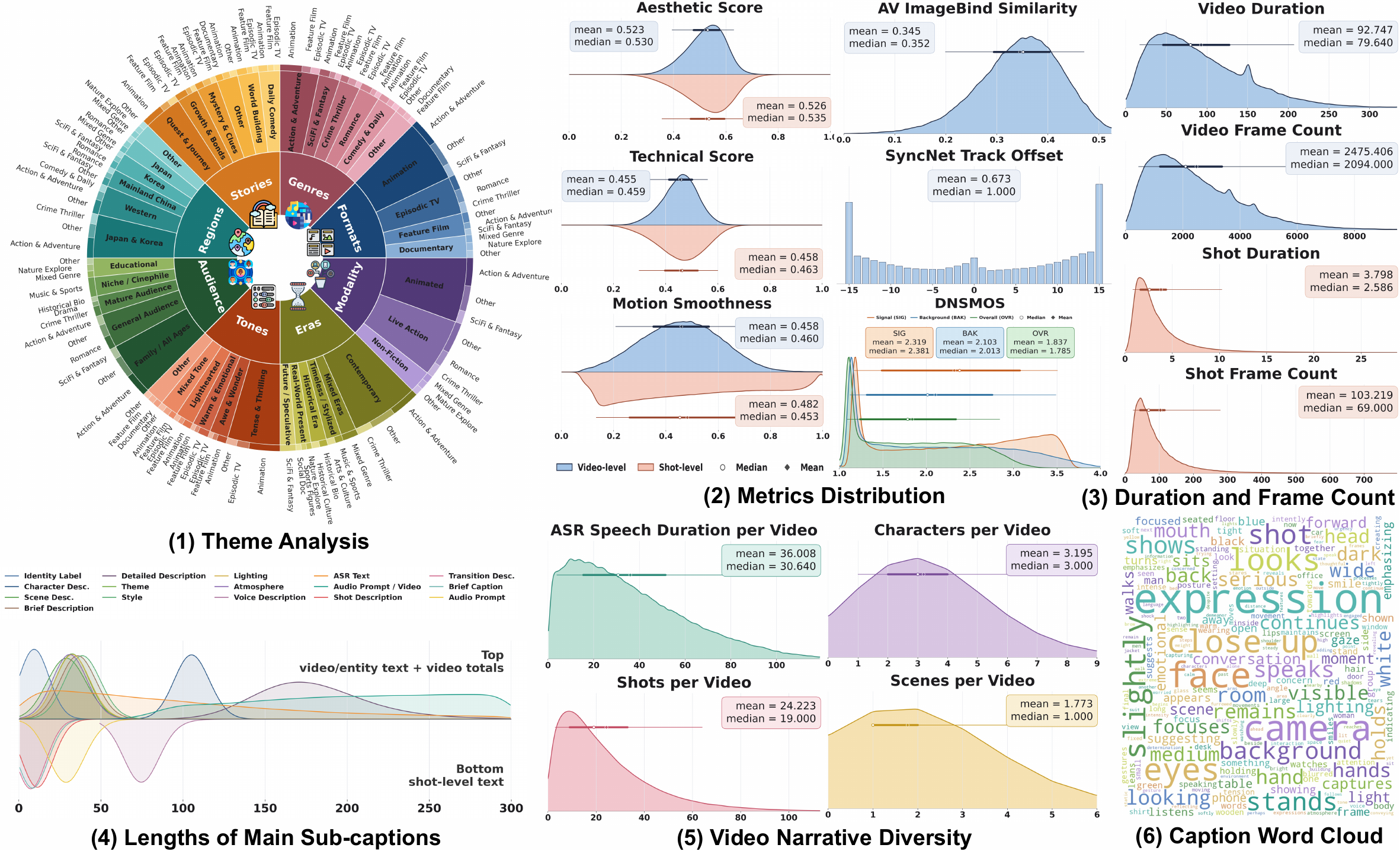}
    \caption{Statistical overview of the \Ours dataset across multiple dimensions.}
    \label{fig:statistic}
\end{figure*}

\noindent\textbf{Windowed audio-video identity binding.}
Given the video clip, audio track, global character anchors, shot-level visual annotations, character voice descriptions, and sentence-level ASR texts, we perform a dedicated cross-modal identity binding step.
It serves two purposes.
\textbf{\textit{1)}} First, it attaches each generated voice description to the correct character anchor $\langle$\texttt{charX}$\rangle$.
\textbf{\textit{2)}} Second, it assigns each ASR sentence to the character who speaks it.
The first task is relatively easier because explicit character-level visual and acoustic cues are already available.
The main challenge lies in resolving the speaker of each sentence in long-form cinematic dialogue, where off-screen speech, shot changes, and overlapping characters frequently introduce ambiguity.
To reduce model hallucinations, we filter out non-speech intervals based on ASR outputs and partition the remaining sequence into localized windows that preserve intact shots and complete spoken sentences.
Given these localized audio-video windows, together with prior visual annotations and ASR texts, the model performs cross-modal binding to assign each sentence to the corresponding character anchor $\langle$\texttt{charX}$\rangle$.
As shown in Tab.~\ref{tab:av_binding_ablation}, this windowed strategy substantially improves ASR segment-to-character binding accuracy.
Compared with 
% the audio-centric speaker-attribution setting in 
Tab.~\ref{tab:speaker_binding_ablation}, 
% the final multimodal binding stage further improves 
Qwen3-Omni-30B-A3B improves from 83.1\% to 95.4\%, indicating that visual grounding provides complementary evidence for binding.
By discarding non-speech intervals and restricting reasoning to localized audio-video contexts, it also reduces computational overhead, making corpus-scale deployment feasible.

% \noindent\textbf{Windowed audio-video identity binding.} To effectively reduce model hallucinations, we actively filter out non-speech intervals based on ASR outputs and partition the remaining sequence into localized windows dynamically sized to encapsulate intact shots and complete spoken sentences. Given these localized audio-video windows, along with prior visual annotations and ASR texts as inputs, the model executes a cross-modal binding step to accurately assign each sentence to the correct character ($\langle$\texttt{charX}$\rangle$). This windowed strategy not only reduces computational overhead but also significantly enhances alignment precision.

\subsection{Statistical Comparison and Analysis}

\noindent\textbf{Comparison with prominent video datasets.} Tab.~\ref{tab:dataset-comparison} comprehensively compares \Ours with other prominent datasets. Notably, \Ours is a pioneering audio-video dataset that extends the average sequence duration to an unprecedented \AvgDur seconds and encompasses over \AvgShots physical shots per sequence. Furthermore, it features highly structured, configurable text prompts for both modalities, providing optimal fine-tuning signals for joint audio-video generation. While recent long-video datasets like MiraData~\cite{miradata} and LVD-2M~\cite{lvd-2m} exist, their low shot counts indicate predominantly single-shot or static single-scene recordings.

\noindent\textbf{Numerical statistics.} Fig.~\ref{fig:statistic} details the statistical distributions of \Ours across six dimensions:
\textbf{\textit{1)}} An eight-dimensional taxonomy (Genre, Format, Region, Modality, Story Logic, Era, Tone, and Audience) ensures high categorical diversity for generalizable generation, constructed with advanced LLM assistance.
\textbf{\textit{2)}} Consistently high scores in video/audio quality and audio-video alignment validate the dataset's overall fidelity, while enabling researchers to set customizable filtering thresholds.
\textbf{\textit{3)}} Duration and frame count distributions at both video and shot levels emphasize the dataset's inherently long-form nature, addressing the critical scarcity of high-quality long-duration priors.
\textbf{\textit{4)}} Prompt length distributions reveal a remarkably dense annotation volume (averaging 6,400 words per video), offering exceptionally fine-grained guidance for controllable synthesis.
\textbf{\textit{5)}} Video-level distributions of shots, characters, unique scenes, and ASR duration strictly quantify the high structural and intricate narrative complexity of our sequences.
\textbf{\textit{6)}} A semantic word cloud highlights the core descriptive and cinematic vocabulary embedded within our annotations.
% Together, these multidimensional statistics highlight the exceptional scale, diversity, and structural richness of \Ours, establishing a robust foundation for future joint generation benchmarks.

\begin{figure*}[t]
    \centering
    \includegraphics[width=\linewidth]{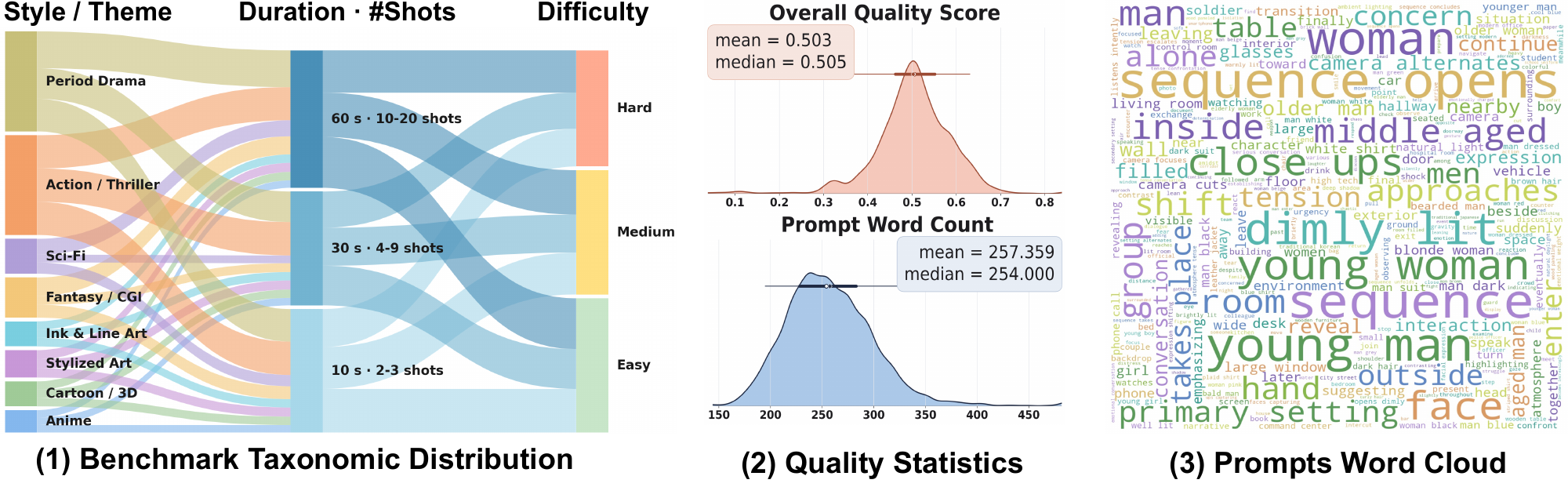}
    \caption{Comprehensive statistical overview of CineBench, illustrating its diverse taxonomic flow, rigorous quality distributions, and rich semantic vocabulary.}
    \label{fig:benchmark}
\end{figure*}

\section{CineBench: Advancing Cinematic Evaluation via Hierarchical Benchmark}
\label{sec3}

To systematically assess model capabilities in complex narrative synthesis, we introduce CineBench, a comprehensive evaluation suite designed for multi-shot, long-form joint audio-video generation.

\subsection{Task Definition}

CineBench evaluates whether a generative model can synthesize a temporally ordered multi-shot sequence from the structured conditions defined in Sec.~\ref{sec23}. 
Each instance follows the same anchor-centric schema as CineDance-1M, but is rendered into model-specific inputs before evaluation. 
The target output is a coherent long-form sequence that preserves the intended characters, scenes, events, and audio-visual content across shots. 
Video-only models are evaluated on visual and narrative dimensions, while native audio-video models are additionally evaluated on audio quality and audio-video synchronization.

\subsection{Benchmark Construction}

\noindent\textbf{Taxonomy and difficulty stratification.}
To comprehensively evaluate multidimensional generative capabilities, we stratify benchmark instances across three dimensions: \textbf{\textit{Theme/Style}}, \textbf{\textit{Duration/Shot Count}}, and \textbf{\textit{Generation Difficulty}}.
Here, \textit{Generation Difficulty} targets the quality challenge within the model's intrinsic capacity boundary, while \textit{Duration/Shot Count} characterizes the capacity boundary itself, i.e., how long a video and how many shots a model can support.
Although current methods are still limited in achievable duration and shot count, this design makes CineBench a forward-looking benchmark that remains meaningful as long-form generation models continue to improve.

The difficulty level is deterministically defined by rule-based features:
\textbf{\textit{1)}} entity complexity, measured by the number of distinct character anchors;
\textbf{\textit{2)}} scene complexity, measured by the number of distinct scene anchors;
and \textbf{\textit{3)}} dialogue/audio complexity, measured by speaker-adjusted ASR length.
Specifically, for each candidate window, we compute
\[
D
=
n_{\mathrm{char}}
+
1.5\,n_{\mathrm{scene}}
+
0.4\log\bigl(1+n_{\mathrm{spk}}L_{\mathrm{ASR}}\bigr),
\]
where $n_{\mathrm{char}}$, $n_{\mathrm{scene}}$, and $n_{\mathrm{spk}}$ denote the numbers of distinct character, scene, and speaker anchors within the window, and $L_{\mathrm{ASR}}$ denotes the cumulative ASR character count.
We consider \textbf{\textit{three}} \textit{Duration/Shot Count} tiers: 
10s with 2--3 shots, 30s with 4--9 shots, and 60s with 10--20 shots.
Within each duration tier, Easy, Medium, and Hard labels are assigned by empirical tertile cuts of $D$, preventing longer windows from being trivially classified as harder.
To ensure a balanced evaluation, we uniformly sample high-quality instances across the resulting \textit{Theme/Style} $\times$ \textit{Duration/Shot Count} $\times$ \textit{Difficulty} grid, ultimately curating a diverse set of \textbf{1000} testing cases.
% More details are provided in Appendix~\ref{app:pipeline-cinebench}.

\noindent\textbf{Prompt and condition construction.} To serve as a universal evaluation suite, \OurBench retains the fine-grained, structured representations of the global and shot-level dual-modal annotations, facilitating seamless adaptation to the diverse input formats required by various methods. Benchmark test cases are carefully sampled from the curated dataset across the aforementioned taxonomy tiers, prioritizing high-quality sequences. 
% For visual conditioning, we initially extract the middle frame of each ground-truth shot. 
Crucially, all sampled textual annotations undergo rigorous manual verification to guarantee absolute correctness and premium visual quality. \textbf{\textit{Finally, to strictly prevent data leakage and ensure fair evaluation, all these test cases are permanently removed from \Ours}}. A comprehensive statistical overview of CineBench is provided in Fig.~\ref{fig:benchmark}.

\subsection{Comprehensive Evaluation Suite}
\label{sec:cinebench_metric}
% \noindent\textbf{Comprehensive evaluation suite.} 
We assess performance across six dimensions using a robust automated suite:
\textbf{\textit{1) Video Quality}} measures shot-level \textit{Aesthetic Quality} (aesthetic-predictor-v2-5)~\cite{vbench}, \textit{Imaging Quality} (MUSIQ)~\cite{musiq}, and \textit{Motion Smoothness} (AMT)~\cite{amt}.
\textbf{\textit{2) Audio Quality}} reports global \textit{AudioBox-Aesthetics} (PQ, CE, CU)~\cite{audiobox} and speech intelligibility measured by WER/CER using Whisper-large-v3.
\textbf{\textit{3) AV Sync}} combines \textit{Sync-C} and \textit{Sync-D}~\cite{syncnet} for overall lip-sync with ImageBind \textit{IB-Score}~\cite{imagebind}.
\textbf{\textit{4) Prompt Alignment}} captures dual-granularity video alignment via \textit{ViCLIP}~\cite{internvid} (shot-level) and \textit{VideoScore-v1.1}~\cite{videoscore} (video-level), while audio alignment utilizes ImageBind \textit{IB-A Score}~\cite{imagebind}.
\textbf{\textit{5) Narrative Continuity}} introduces annotation-grounded alternatives to address the critical failure of standard frame-level metrics that inappropriately penalize legitimate camera cuts.
Specifically, \textit{Identity Continuity} leverages ArcFace~\cite{arcface} clustering matched against declared $\langle\text{char}_k\rangle$ tokens to score cross-shot character consistency, and \textit{Scene Continuity} computes the mean pairwise DINOv2~\cite{dinov2} cosine across shots sharing the same $\langle\text{scene}_k\rangle$ token.
\textbf{\textit{6) Shot Structure Response}} evaluates whether the generated video responds to the target multi-shot structure by comparing the ground-truth shot partition $\{I_i\}_{i=1}^{N}$ with the detected generated shot partition $\{\hat I_j\}_{j=1}^{M}$.
We compute
\[
\begin{array}{@{}c@{\;}c@{}}
\displaystyle
S_{\mathrm{cnt}}
=
\tfrac{\min(N,M)}{\max(N,M)}
&
\begin{aligned}
S_{\mathrm{seg}}
&=
\tfrac{1}{2}
\left(
\tfrac{1}{N}\sum_i \max_j \mathrm{IoU}(I_i,\hat I_j)
\right.\\[-1pt]
&\left.
\quad+
\tfrac{1}{M}\sum_j \max_i \mathrm{IoU}(I_i,\hat I_j)
\right).
\end{aligned}
\end{array}
\]
Here, $S_{\mathrm{cnt}}$ measures shot-count agreement, while $S_{\mathrm{seg}}$ measures bidirectional temporal overlap between the target and generated shot partitions.
The final Shot Structure Response is
$
\mathrm{SSR}
=
S_{\mathrm{cnt}}^{0.35}
S_{\mathrm{seg}}^{0.65}.
$
% Further metric implementation details are available in Appendix~\ref{sec:suppl:cinebench-metric-impl}.

\subsection{Human Validation Protocol}

To assess whether CineBench automatic scores reflect human perception, we conduct a randomized and double-blind human study on sampled generated sequences. 
Each generated video is rated by 10 independent evaluators using a 5-point Likert scale (1 = unusable, 2 = poor, 3 = acceptable, 4 = good, 5 = excellent), with all source models anonymized and presentation order randomized to reduce subjective bias. 
The scoring rubric follows the same six dimensions as our automatic suite: 
\textbf{\textit{1)}} \textit{Video Quality} evaluates per-frame fidelity, visual artifacts, and motion smoothness; 
\textbf{\textit{2)}} \textit{Audio Quality} measures speech clarity, background sound naturalness, and the absence of audible artifacts; 
\textbf{\textit{3)}} \textit{Audio-Video Synchronization} assesses lip-sync precision for speech segments and semantic alignment between sound events and visual content; 
\textbf{\textit{4)}} \textit{Prompt Alignment} measures faithfulness to the rendered benchmark condition, including characters, scenes, events, dialogue, and sound descriptions; 
\textbf{\textit{5)}} \textit{Narrative Continuity} evaluates cross-shot identity preservation, scene recurrence, object persistence, and ordered event progression; 
and \textbf{\textit{6)}} \textit{Shot Structure Response} evaluates whether the generated video exhibits the intended number of shots, clear shot transitions, and a temporal shot layout consistent with the target structure.
We use inter-rater agreement to measure panel consistency and Spearman rank correlation to quantify the alignment between automatic CineBench scores and human judgments. 
Detailed results analyses are reported in Sec.~\ref{sec5}.

\section{CineDance: Multi-Shot Long-Form Audio-Video Generation}
\label{sec4}

% \noindent\textbf{Backbone.} We select the state-of-the-art LTX-2.3~\cite{hacohen2026ltx} as our backbone strictly for its native capability in joint audio-video Generation, which perfectly aligns with the dual-modal nature of CineDance.

\subsection{Backbone and Joint Audio-Video Formulation}
\label{sec:method_backbone}

\noindent\textbf{Backbone.}
We select the state-of-the-art LTX-2.3~\cite{ltx-2} as our backbone strictly for its native capability in joint audio-video generation, which naturally aligns with the dual-modal nature of CineDance.
% LTX-2.3 is a DiT-based audio-video foundation model.
% that generates synchronized visual and audio signals within a unified generative framework.
Its core design is an asymmetric dual-stream DiT, where a 13B-parameter video branch and a 3B-parameter audio branch process modality-specific latent tokens.
The two streams are bridged by a 3B-parameter audio-video cross-attention branch, allowing temporally aligned information exchange between visual and acoustic representations.

\noindent\textbf{Joint audio-video formulation.}
Given a structured textual condition $c$, video latent $z_v$, and audio latent $z_a$, the backbone learns a conditional joint generative model
$
p_\theta(z_v,z_a\mid c),
$
through a modality-aware latent denoising objective:
$
\mathcal{L}_{\mathrm{AV}}
=
\mathbb{E}_{t}
\left[
\lambda_v
\mathcal{L}_v(z_{v,t},t,c)
+
\lambda_a
\mathcal{L}_a(z_{a,t},t,c)
\right],
$
where $\mathcal{L}_v$ and $\mathcal{L}_a$ denote the video and audio denoising losses, respectively.
The coefficients $\lambda_v$ and $\lambda_a$ control the relative contribution of the two modality-specific objectives.
% In our experiments, we use a balanced weighting between the two modalities, i.e., $\lambda_v:\lambda_a=1:1$.

\subsection{Rethinking Multi-Shot Long-Form Generation: A Conditioning and Output Perspective}
\label{sec:rethinking}
As surveyed in Sec.~\ref{sec:rw3}, the multi-shot generation landscape has been predominantly described through the lens of methodological mechanisms, such as masked attention, autoregressive frameworks, and sparse attention patterns. 
While this mechanism-centric taxonomy reflects the methodological evolution of existing approaches, it remains a descriptive categorization of how methods operate internally. 
In this section, we step back and propose a more fundamental taxonomy at a higher level of abstraction, organized by \textbf{\textit{input conditions}} and \textbf{\textit{output formats}}.
We group existing methods into the following \textbf{\textit{three}} paradigms:
\textbf{\textit{1)}} \textbf{\textit{Per-shot generation with memory propagation.}}
This paradigm decomposes multi-shot long-form generation into separate short-video generation tasks and stitches the resulting clips afterward~\cite{stage,storymem,vgot,movieagent,onestory}. 
Cross-shot consistency is mainly maintained through memory propagation, such as carrying latent frames from previous shots into later generations. 
Keyframe-based methods also fall into this category, since keyframes usually serve as the start and end frames of individual short clips, and the coherence of these keyframes largely determines the coherence of the assembled multi-shot video. 
This per-shot decomposition tends to produce shots with similar or fixed durations, which may conflict with the desired narrative rhythm and temporal logic. 
Although memory propagation can mitigate drift, cross-shot consistency remains limited because different shots are still generated separately.
\textbf{\textit{2)}} \textbf{\textit{Structure-aware single-forward generation.}}
This paradigm requires detailed multi-shot structure as input, including the start and end times of each shot, shot-level prompts, and sometimes finer control signals~\cite{holocine,cinetrans,mask2dit,multishotmaster}. 
The model generates the full video in a single forward process, using shot-aware attention masks, segment-wise prompt assignment, or RoPE-based temporal indexing to separate and align shots. 
While this avoids post-hoc stitching and provides explicit layout control, it relies heavily on user-specified structure. 
Shot durations cannot adapt freely to narrative rhythm, imperfect conditions may introduce artifacts, and attention manipulation can still leave visible cross-shot consistency errors.
\textbf{\textit{3)}} \textbf{\textit{Reference-to-multi-shot generation.}}
Recent closed or commercial systems, such as Seedance~2.0~\cite{seedance2} and SkyReels-V4~\cite{skyreels-v4}, introduce an emerging reference-to-multi-shot setting, where the reference image is arranged as a grid of shot-level visual guidance. 
Importantly, the sub-images in the grid are not start or end frames to be reproduced exactly; they provide high-level visual-semantic guidance for scene, style, and composition. 
The model interprets these visual cues and automatically organizes the output into multiple shots, without requiring user-specified shot boundaries. 
Compared with the previous paradigms, this setting allows shot durations to adapt more naturally to the generated narrative and often yields stronger visual consistency in our benchmarking observations. 
While this paradigm is flexible and user-friendly, it remains less explored in open-source methods.
\textbf{\textit{Overall}}, these paradigms still rely on explicit structural aids, ranging from user-specified shot boundaries to reference-based visual guidance.
% , with the reference-to-multi-shot paradigm further enabling adaptive shot organization from visual references 
% CineDance aims to support end-to-end adaptive multi-shot generation while reducing the need for manually specified shot schedules or mandatory references at inference.

\begin{figure}[t]
    \centering
    \includegraphics[width=\linewidth]{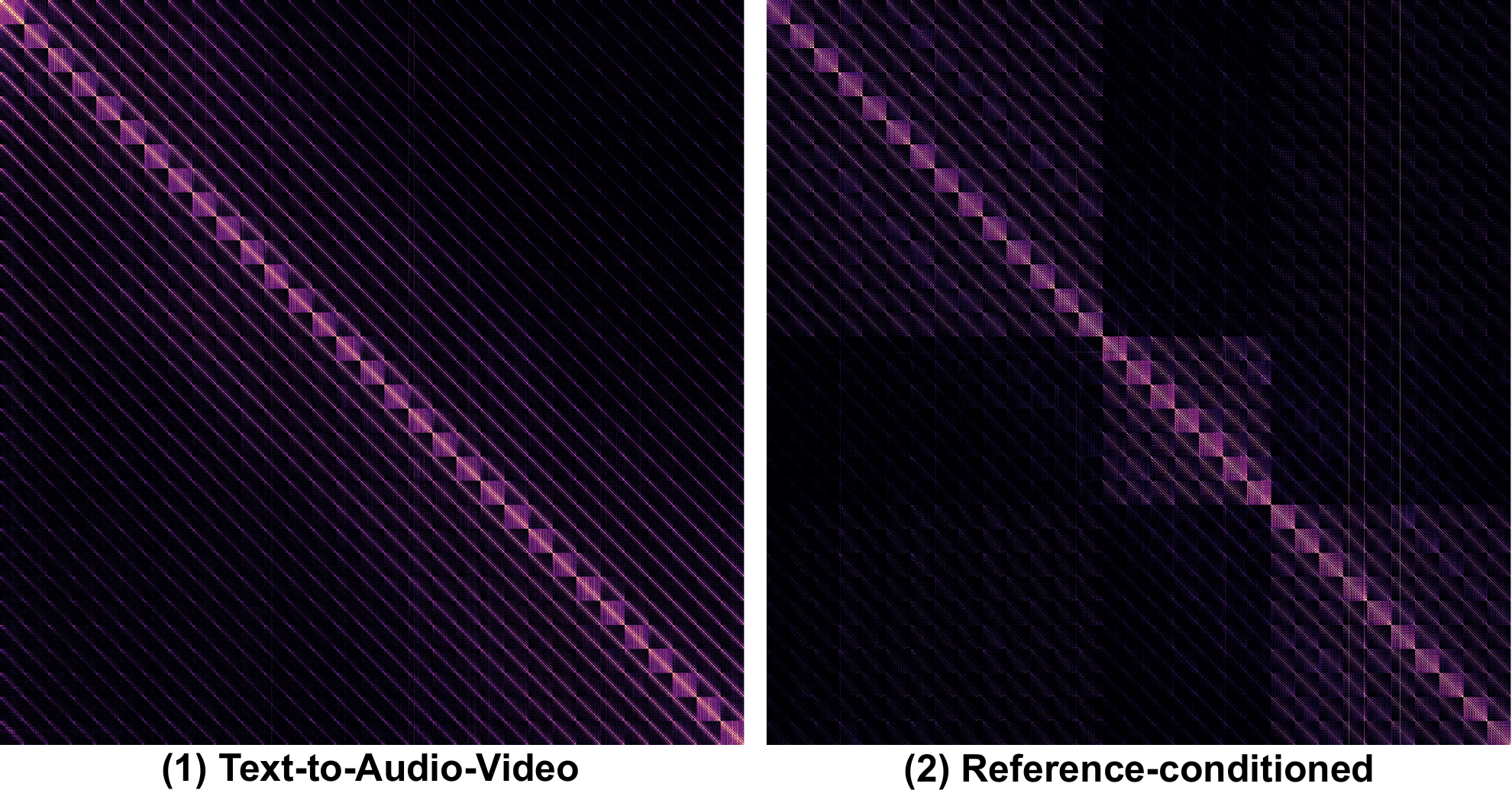}
\caption{
Self-attention map visualization under the same generation condition. 
Both maps are extracted from the same transformer block at the same intermediate denoising step, averaged over attention heads. 
\textbf{(1)} Direct T2AV generation mainly exhibits shot-agnostic temporal locality, \textbf{(2)} whereas reference conditioning produces clearer segment-level attention structure. 
}
    \label{fig:selfattn}
\end{figure}

\subsection{Reference Frames as Training-Time Scaffolds}
\label{sec:method_scaffold}

\noindent\textbf{Observation and Motivation.} A direct way to adapt a native T2AV backbone to multi-shot generation is to train it with structured shot-level prompts. 
However, this strategy provides only sparse textual supervision for a highly structured generation problem. 
The model must simultaneously infer shot boundaries and durations, preserve cross-shot visual continuity, and maintain audio-video consistency from text alone, making direct text-conditioned adaptation unstable and inefficient.
This difficulty is consistent with the gap observed in Sec.~\ref{sec:intro}, where existing foundation models respond weakly to shot-structured prompts and often fail to form clear multi-shot organization. 
We further diagnose this limitation through temporal attention visualization. 
As shown in Fig.~\ref{fig:selfattn}, the direct T2AV baseline exhibits a monotonically decaying \textbf{\textit{temporal locality pattern}} that is agnostic to shot boundaries~\cite{attentionanalysis}. 
Its attention mass falls off smoothly with frame distance and shows no discontinuity at the prompt-specified shot boundaries. 
This suggests that the shot organization specified in the prompt is not effectively reflected in the backbone's temporal attention.
Motivated by the adaptive shot organization enabled by \textit{reference-to-multi-shot generation}, CineDance uses reference frames as \textbf{\textit{training-time visual-temporal scaffolds}} rather than mandatory inference-time inputs. 
We first make multi-shot organization easier to learn with dense visual and temporal anchors, and then gradually weaken these anchors so that the model can retain the learned structure under reduced inference conditions.

\noindent\textbf{Training-time visual-temporal scaffolds.}
Since the internal mechanisms of recent reference-to-multi-shot systems are not publicly specified, we implement a lightweight yet effective reference-conditioning interface following common practice in open reference-conditioned video generation.
Specifically, each reference frame is encoded by the video VAE into a latent representation and appended to the video latent tokens as additional reference tokens, serving as the \textbf{\textit{visual scaffold}} that provides dense cues for scene layout, style, composition, and recurring visual content.

A key design choice is to assign proper RoPE~\cite{rope} indices to these reference tokens.
Each reference latent uses the same spatial RoPE indices as a normal video latent frame, so that the reference content, together with its spatial RoPE assignment, can be aligned with the spatial layout of generated video tokens.
LTX-2 uses the timestamp of each conditioning frame, computed from its frame index and the target frame rate, as its temporal RoPE index.
To make the initial training well-conditioned, we assign each reference frame the ground-truth temporal index of its corresponding frame in the training video, forming a simple but effective \textbf{\textit{temporal scaffold}}.
% that indicates where the reference appears in the sequence and makes early multi-shot organization easier to learn.

We formalize the reference scaffold as
$s_{\mathrm{ref}}(\eta_v,\eta_t)
=
\{(\rho(r_k;\eta_v),\psi(\tau_k;\eta_t))\}_{k=1}^{K}$,
where $r_k=\mathcal{E}_{\mathrm{VAE}}(x^{\mathrm{ref}}_k)$ is the VAE latent of the $k$-th reference frame, and $\tau_k$ is its ground-truth frame-level timestamp.
We further introduce two strength parameters, $\eta_v$ and $\eta_t$, to control the guidance strength of the visual and temporal scaffolds, respectively.

\noindent\textbf{Progressive scaffold removal.}
The training-time scaffold should make optimization easier, but it should not remain a mandatory inference-time input. 
We therefore progressively weaken and remove the reference scaffold through the \textbf{\textit{D}}ual-\textbf{\textit{A}}xis \textbf{\textit{R}}eference \textbf{\textit{C}}urriculum (\textbf{\textit{DARC}}), a two-step reference-strength annealing process.

\textbf{\textit{For visual scaffold}},
We first weaken the visual scaffold through continuous noising of the reference latent. 
Given the clean reference latent $r_k$, we define
\[
\rho(r_k;\eta_v(u))
=
\eta_v(u) r_k
+
\bigl(1-\eta_v(u)\bigr)\epsilon_k,
\quad
\epsilon_k\sim\mathcal{N}(0,I),
\]
where $u$ denotes the training step and $\eta_v(u)\in[0,1]$ is a \textbf{\textit{monotonically decreasing}} visual-reference strength schedule. 
As $\eta_v(u)$ decreases, the reference latent is gradually interpolated from a clean visual anchor to pure Gaussian noise.

Even when the reference latent is fully noised, the model may still rely on the temporal RoPE index of the reference token to organize shot placement and duration.
\textbf{\textit{For the temporal scaffold}}, we therefore relax the reference token's temporal index through a stochastic switch:
\[
\psi(\tau_k;\eta_t)=
\begin{cases}
\tau_k, & \text{with probability } 1-q(\eta_t),\\
\hat{\tau}_k, & \text{with probability } q(\eta_t),
\end{cases}
\]
where $\tau_k$ is the ground-truth timestamp of the corresponding reference frame, and $\hat{\tau}_k=\lfloor kN/(K+1)\rfloor$ is a placement-independent ordinal index.
The switching probability $q(\eta_t)$ increases as the temporal strength decreases, so the reference tokens retain only their relative temporal order while losing their exact timestamps.

Although the reference latent and its RoPE assignment no longer provide reliable visual details or exact timestamp information, the reference token itself remains in the sequence and may still act as an extra implicit conditioning slot, hindering the training condition from matching the true T2AV regime.
We therefore complete the scaffold removal with a \textbf{\textit{final reference-dropping stage}}, where the reference latents and their RoPE indices are removed together:
\[
\bar{s}_{\mathrm{ref}} =
\begin{cases}
s_{\mathrm{ref}}, & \text{with probability } 1-p_{\mathrm{drop}}(u),\\
\varnothing, & \text{with probability } p_{\mathrm{drop}}(u),
\end{cases}
\]
where $\varnothing$ removes all reference tokens and their positional assignments from the input sequence.
The dropping probability $p_{\mathrm{drop}}(u)$ increases in the late stage of training, gradually turning weakened reference conditioning into genuine text-only T2AV training.

\begin{figure}[t]
    \centering
    \includegraphics[width=\linewidth]{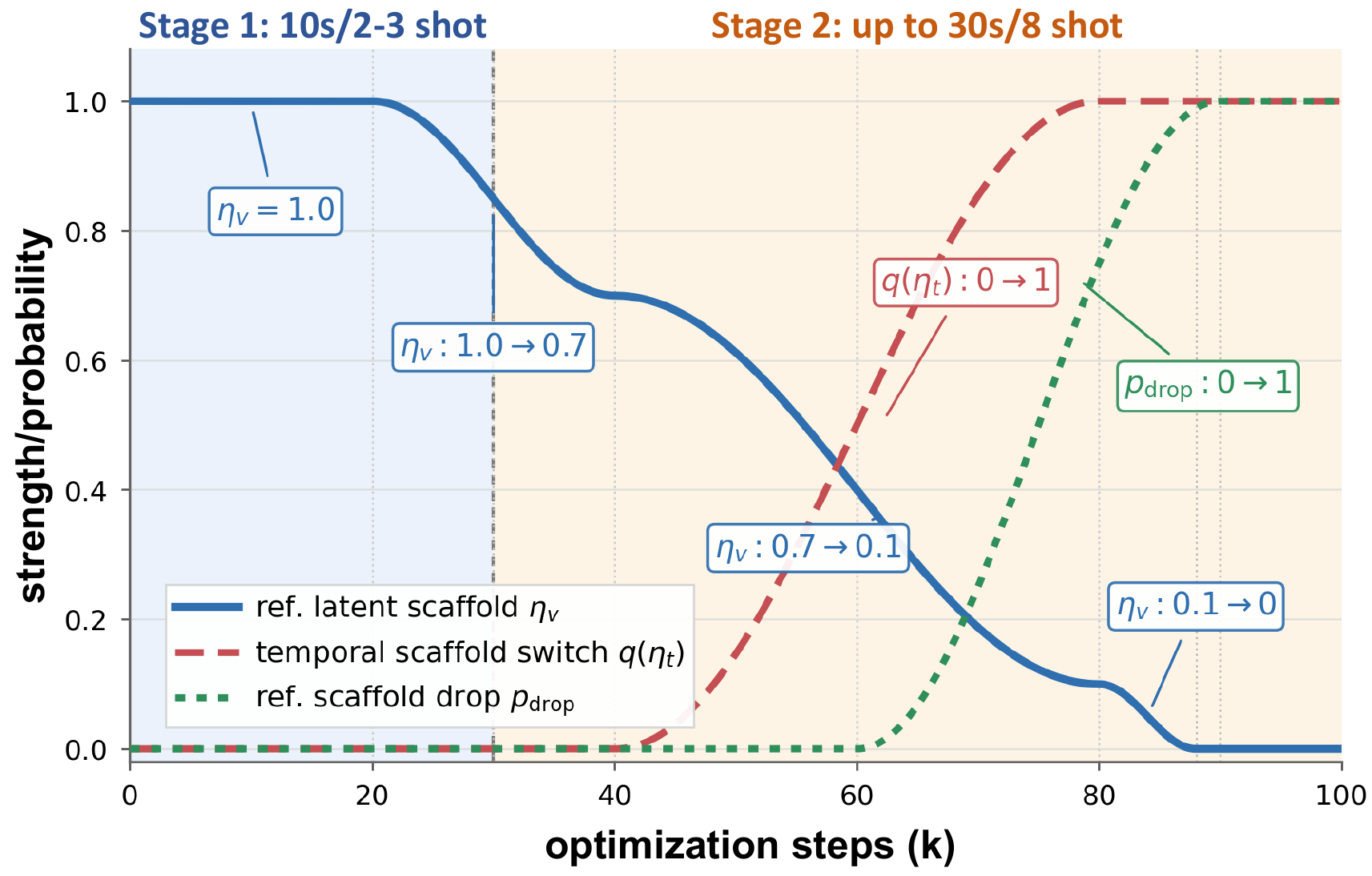}
\caption{
Overall training schedule.
\textbf{\textit{1)}} The background color shows the data-driven curriculum.
\textbf{\textit{2)}} The three curves summarize DARC: the visual scaffold strength $\eta_v$ decreases, the temporal-switch probability $q(\eta_t)$ increases, and the reference-dropping probability $p_{\mathrm{drop}}$ is activated in the late stage.
}
    \label{fig:schedule}
\end{figure}

\begin{figure*}[t]
    \centering
    \includegraphics[width=\linewidth]{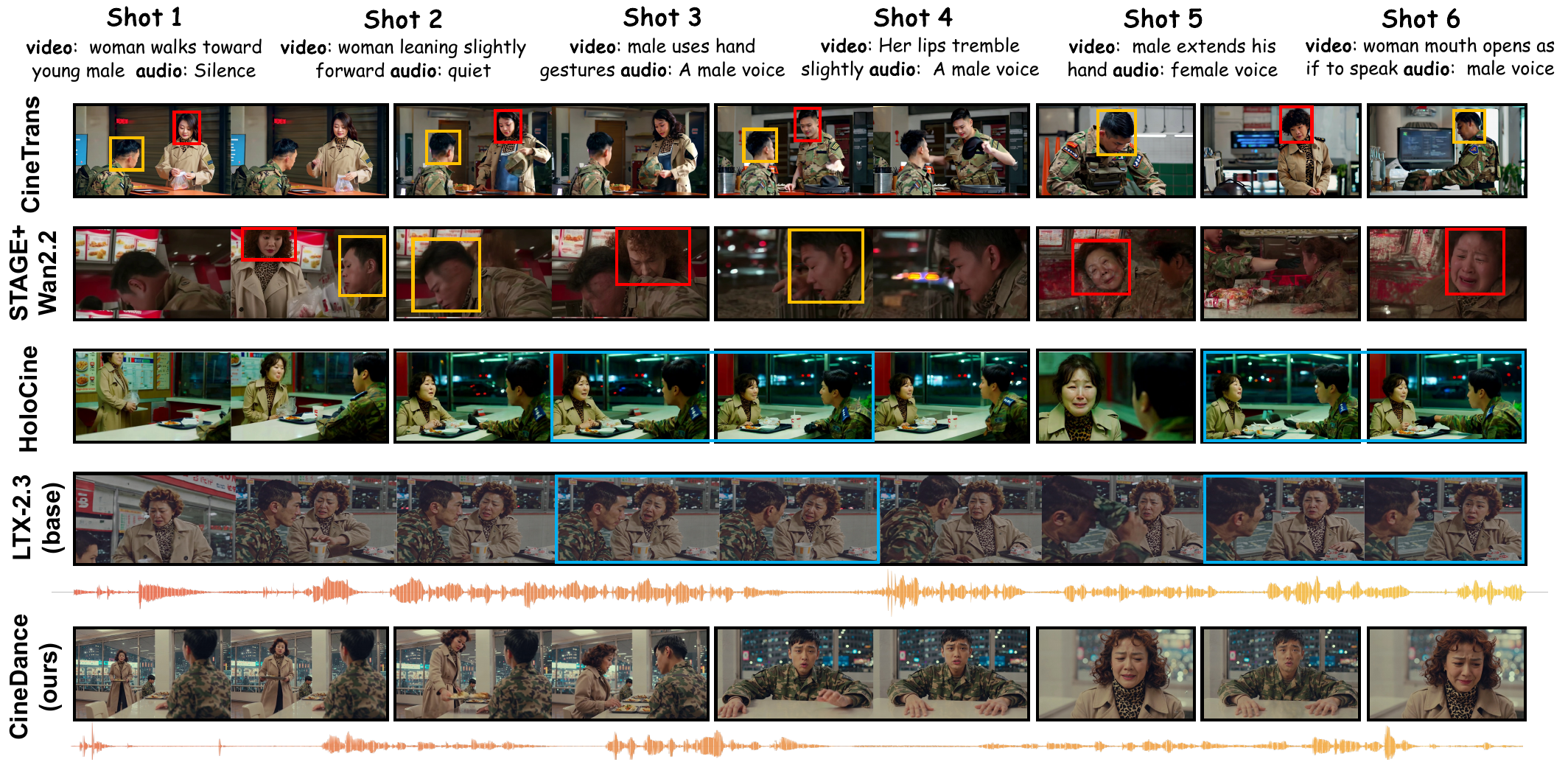}
    \caption{Qualitative comparison of CineDance with baseline models. \textbf{Red and yellow boxes} indicate character identity inconsistencies across shots, while \textbf{blue boxes} denote failed shot transitions.}
    \label{fig:compare}
\end{figure*}

% \subsection{Training Recipe and Implementation Details}
\subsection{Long-Form Adaptation and Implementation Details}
\label{sec:method_impl}
\noindent\textbf{Data-driven curriculum.}
The previously introduced DARC mainly addresses the latent-domain gap when adapting the backbone from single-shot generation to multi-shot generation.
Since CineDance further targets long-form audio-video generation, we additionally introduce a data-driven curriculum to expand the model's effective temporal capacity:
\textbf{\textit{1)}} Stage~1 trains on short windows of 2--3 adjacent shots with durations of 10--12 seconds, so that the model first learns local shot switching.
\textbf{\textit{2)}} Stage~2 extends the training window to up to 8 shots and approximately 30 seconds, encouraging longer-range cross-shot consistency.
Both stages sample temporally contiguous windowed video clips from the original long-form sequences within CineDance-1M.

\noindent\textbf{Structured prompt organization.}
For each windowed video clip containing $M$ selected shots, we render a structured textual condition consisting of character and scene headers followed by re-indexed shot-level blocks:
\[
c_{\mathrm{text}}
=
\big[
\underbrace{\mathcal{H}_{\mathrm{char}},\mathcal{H}_{\mathrm{scene}}}_{\text{global headers}},
\underbrace{\mathcal{B}_1,\ldots,\mathcal{B}_M}_{\text{shot blocks}}
\big].
\]
The headers are defined as
\[
\begin{aligned}
\mathcal{H}_{\mathrm{char}}
&=
\{\,\langle\mathrm{char}c\rangle=e_c\,\}_{c=1}^{C},\\
\mathcal{H}_{\mathrm{scene}}
&=
\{\,\langle\mathrm{scene}s\rangle=g_s\,\}_{s=1}^{S},
\end{aligned}
\]
where $e_c$ and $g_s$ denote the character descriptions and scene description, respectively.
\textit{Only characters and scenes appearing in the selected shots are included.}
Each shot block is rendered as
\[
\begin{aligned}
\mathcal{B}_i
=
&\big[
\mathrm{SHOT}\ i
\mid
\mathrm{scene}\ s_i
\mid
\mathrm{camera}\ \kappa_i
\big]
\oplus d_i^{v}
\oplus d_i^{a} \\
&\oplus
\{(\mathrm{spk}_{i,\ell},\mathrm{speech}_{i,\ell})\}_{\ell=1}^{L_i},
\end{aligned}
\]
where $s_i$ is the scene anchor, $\kappa_i$ is the camera descriptor, $d_i^{v}$ and $d_i^{a}$ are the visual and audio descriptions, and $(\mathrm{spk}_{i,\ell},\mathrm{speech}_{i,\ell})$ denotes the $\ell$-th ASR segment in shot $i$, with a speaker anchor and the corresponding transcript.
Regardless of the original shot indices in the source sequence, the selected shots are re-indexed from $1$ to $M$ to provide a compact local timeline.

\begin{table*}[t]
    \centering
    \caption{
Quantitative comparison of \OurModel with SOTA baselines on \OurBench. Best and second-best results are \textbf{bolded} and \underline{underlined}. The ``--'' symbol denotes no audio generation capability. \Omark~indicates original implementations, while \Lmark~utilizes LTX-2.3 as the video generator.
    }
    \label{tab:main-results}
    \setlength{\tabcolsep}{3.5pt}
    \resizebox{\linewidth}{!}{
        \renewcommand{\arraystretch}{1.15}
        \begin{tabular}{@{}l|ccc|cccc|cc|ccc|cc@{}}
            \toprule
            \multirow{2}{*}{\textbf{Method}} & \multicolumn{3}{c|}{\textbf{Video Quality}} & \multicolumn{4}{c|}{\textbf{Audio Quality}} & \multicolumn{2}{c|}{\textbf{AV Sync}} & \multicolumn{3}{c|}{\textbf{Prompt Alignment}} & \multicolumn{2}{c}{\textbf{Continuity}} \\
            \cmidrule(lr){2-4} \cmidrule(lr){5-8} \cmidrule(lr){9-10} \cmidrule(lr){11-13} \cmidrule(l){14-15}
            & AQ $\uparrow$ & IQ $\uparrow$ & MS $\uparrow$ & PQ $\uparrow$ & CE $\uparrow$ & CU $\uparrow$ & WER $\downarrow$ & Sync-C$\uparrow$/D$\downarrow$ & IB-S $\uparrow$ & ViC $\uparrow$ & VSc $\uparrow$ & IB-A $\uparrow$ & ID $\uparrow$ & Scene $\uparrow$ \\
            \midrule
            CineTrans~\cite{cinetrans} & 0.47 & 0.48 & 0.97 & -- & -- & -- & -- & -- & -- & 0.14 & 2.92 & -- & 0.21 & \underline{0.63} \\
            Mask2DiT~\cite{mask2dit} & 0.56 & \underline{0.67} & \textbf{0.99} & -- & -- & -- & -- & -- & -- & 0.14 & 2.76 & -- & 0.23 & 0.55 \\
            MultiShotMaster~\cite{multishotmaster} & 0.51 & 0.51 & 0.96 & -- & -- & -- & -- & -- & -- & 0.15 & 3.19 & -- & 0.13 & 0.37 \\
            HoloCine~\cite{holocine} & 0.50 & 0.36 & \textbf{0.99} & -- & -- & -- & -- & -- & -- & 0.17 & \underline{3.21} & -- & 0.12 & 0.58 \\
            % \midrule
            StoryMem~\cite{storymem} & 0.57 & 0.66 & \underline{0.98} & -- & -- & -- & -- & -- & -- & 0.10 & 2.68 & -- & \underline{0.26} & 0.53 \\
            VGoT~\cite{vgot} \Omark & 0.42 & 0.53 & 0.91 & -- & -- & -- & -- & -- & -- & 0.19 & 0.72 & -- & 0.18 & 0.52 \\
            VGoT~\cite{vgot} \Lmark & 0.45 & 0.55 & 0.93 & 6.45 & 4.05 & 5.75 & 0.73 & \underline{1.42} / 9.45 & 0.18 & 0.14 & 0.75 & 0.11 & 0.16 & 0.49 \\
            MovieAgent~\cite{movieagent} \Omark & 0.54 & 0.62 & 0.95 & -- & -- & -- & -- & -- & -- & 0.20 & 0.71 & -- & 0.15 & 0.54 \\
            MovieAgent~\cite{movieagent} \Lmark & \underline{0.58} & 0.65 & 0.97 & 6.58 & \underline{4.12} & 5.82 & 0.69 & 1.35 / 9.12 & 0.21 & 0.15 & 0.78 & 0.13 & 0.21 & 0.57 \\
            STAGE~\cite{stage} \Omark & 0.49 & 0.50 & \underline{0.98} & -- & -- & -- & -- & -- & -- & \textbf{0.23} & 2.82 & -- & 0.03 & 0.35 \\
            STAGE~\cite{stage} \Lmark & 0.42 & 0.44 & 0.92 & 6.32 & 3.65 & 5.50 & 0.81 & 1.28 / 9.88 & 0.15 & 0.16 & 0.65 & 0.10 & 0.12 & 0.39 \\
            LTX-2.3 (Base)~\cite{ltx-2} & 0.52 & 0.56 & \underline{0.98} & \underline{6.66} & \textbf{4.20} & \textbf{5.93} & \underline{0.66} & 1.11 / \textbf{8.42} & \underline{0.26} & 0.05 & 3.04 & \underline{0.17} & 0.23 & \underline{0.63} \\
            \midrule
            \rowcolor{TableAccent}
            \textbf{CineDance (Ours)} & \textbf{0.59} & \textbf{0.68} & \textbf{0.99} & \textbf{6.72} & \textbf{4.20} & \underline{5.91} & \textbf{0.52} & \textbf{1.53} / \underline{8.46} & \textbf{0.27} & \underline{0.21} & \textbf{3.22} & \textbf{0.19} & \textbf{0.30} & \textbf{0.69} \\
            % \rowcolor{TableAccent}
            % \textbf{CineDance-KF (Ours)} & \textbf{0.72} & \underline{0.69} & 0.97 & \textbf{7.30} & \textbf{5.45} & \underline{6.41} & \textbf{4.82} / \textbf{\hphantom{0}8.45} & \underline{0.19} & \textbf{0.32} & \textbf{0.82} & \textbf{0.15} & \textbf{0.98} & \textbf{0.96} \\
            \bottomrule
        \end{tabular}
    }
    % \vspace{-10pt}
\end{table*}

\noindent\textbf{Training details.}
Fig.~\ref{fig:schedule} summarizes the overall training schedule, including all curriculum and scaffold-removal stages introduced above.
To ensure smooth transitions across training phases, all curriculum control variables are updated with cosine schedules.
We initialize the model from LTX-2.3 and fully fine-tune it using AdamW~\cite{adamw} for 100K optimization steps, with a global batch size of 32, a constant learning rate of $5\times10^{-5}$, and a weight decay of $0.01$.
All videos are trained at 480p resolution, resized to $480\times832$, with a frame rate of 24 FPS; audio is resampled to 16 kHz.
We train in bfloat16 precision and use a balanced audio-video loss weighting $\lambda_v:\lambda_a=1:1$.

\section{Experiments}
\label{sec5}

\subsection{Experimental Setup}
We evaluate all methods on \OurBench using the automatic metric suite defined in Sec.~\ref{sec:cinebench_metric}.
Although \OurBench includes three \textit{Duration/Shot Count} tiers, existing methods still have limited support for stable 60s audio-video generation.
Therefore, we conduct the main benchmark on \textbf{\textit{1)}} 10s with 2--3 shots, \textbf{\textit{2)}} 30s with 4--9 shots two tiers, which cover both short local shot transitions and longer multi-shot consistency while remaining feasible for all compared methods.
The 60s with 10--20 shots tier is retained in \OurBench as a forward-looking evaluation split and is reported separately when a method supports such generation.

\subsection{Main Comparison}
% \noindent\textbf{Comparison and analysis.} 
\noindent\textbf{Baselines.}
Following the taxonomy in Sec.~\ref{sec:rethinking}, we compare \OurModel with representative methods from two existing multi-shot generation paradigms.
\textbf{\textit{1)}} For per-shot generation with memory propagation, we include STAGE~\cite{stage}, VGoT~\cite{vgot}, MovieAgent~\cite{movieagent}, and StoryMem~\cite{storymem}.
To ensure fair comparison under a unified generator capacity, we additionally evaluate STAGE, VGoT, and MovieAgent after standardizing their underlying video generator to LTX-2.3.
StoryMem involves its own fine-tuning stage, and therefore we do not construct an additional LTX-2.3-based variant for it.
\textbf{\textit{2)}} For structure-aware single-forward generation, we include CineTrans~\cite{cinetrans}, Mask2DiT~\cite{mask2dit}, MultiShotMaster~\cite{multishotmaster}, and HoloCine~\cite{holocine}.
We also include the base LTX-2.3 model~\cite{ltx-2} as the backbone baseline to measure the effect of CineDance adaptation.
For all methods, we convert the structured annotations in CineBench into the \textit{most faithful input conditions} supported by each method, including shot-level prompts or shot structures.
All methods are evaluated at 480p resolution, which falls within the supported resolution range of all baselines; for frame rate, we follow the recommended settings in each official implementation.
\noindent\textbf{Qualitative Comparison.}
As shown in Fig.~\ref{fig:compare}, different baseline paradigms exhibit distinct failure modes under complex multi-shot narratives.
Specifically, CineTrans and STAGE suffer from severe identity inconsistency and video quality degradation.
HoloCine maintains moderate background consistency, but occasionally shows weak shot response and fails to produce clear transitions aligned with the target structure.
Meanwhile, the base LTX-2.3 fails to follow multi-shot textual conditions and exhibits spatial degradation over extended durations.
In contrast, \OurModel intrinsically acquires shot-transition capability without explicit condition injection, preserving the base model's strong visual fidelity while aligning more faithfully with structural prompts. \\

\begin{table}[t]
\centering
\caption{
Shot Structure Response on methods without oracle shot-layout control.
}
\label{tab:ssr}
\begin{tabular}{lccc}
\toprule
Method & $S_{\mathrm{cnt}}\uparrow$ & $S_{\mathrm{seg}}\uparrow$ & SSR $\uparrow$ \\
\midrule
STAGE
& \multirow{4}{*}{1.0000}
& \multirow{4}{*}{0.5220}
& \multirow{4}{*}{0.6553} \\
StoryMem & & & \\
VGoT & & & \\
MovieAgent & & & \\
LTX-2.3 (Base) & 0.6682 & 0.3821 & 0.4646 \\
\midrule
\rowcolor{TableAccent}
\OurModel (ours) & 0.9657 & 0.5866  & 0.6985 \\
\bottomrule
\end{tabular}
\end{table}

\begin{table*}[t]
\centering
\caption{
Ablation studies on CineBench.
The upper block analyzes the effect of training data, while the lower block compares training strategies under the same CineDance-1M setting.
CD-1M denotes CineDance-1M. Best results are \textbf{bolded}.
}
\label{tab:ablation}
\begin{tabular}{@{}l|cc|ccc|cc|c@{}}
\toprule
\multirow{2}{*}{\textbf{Ablated Variants}}
& \multicolumn{2}{c|}{\textbf{AV Sync}}
& \multicolumn{3}{c|}{\textbf{Prompt Alignment}}
& \multicolumn{2}{c|}{\textbf{Continuity}}
& \multirow{2}{*}{\textbf{SSR} $\uparrow$} \\
\cmidrule(lr){2-3}
\cmidrule(lr){4-6}
\cmidrule(lr){7-8}
& Sync-C $\uparrow$/D $\downarrow$
& IB-S $\uparrow$
& ViC $\uparrow$
& VSc $\uparrow$
& IB-A $\uparrow$
& ID $\uparrow$
& Scene $\uparrow$
& \\
\midrule
\multicolumn{9}{@{}l}{\textit{Effect of training data}} \\
OpenhumanVid (1M clips)
& 1.08 / 8.56 & 0.24 & 0.06 & 3.02 & 0.15 & 0.22 & 0.60 & 0.4412 \\
OpenhumanVid (full)
& 1.15 / 8.38 & 0.26 & 0.08 & 3.06 & 0.17 & 0.24 & 0.63 & 0.4785 \\
CD-1M shot-only clips
& 1.28 / 8.92 & 0.23 & 0.14 & 3.12 & 0.16 & 0.19 & 0.55 & 0.5246 \\
CD-1M w/o structured shot-level ann.
& 1.41 / 8.62 & 0.25 & 0.17 & 3.15 & 0.18 & 0.27 & 0.66 & 0.6215 \\
\rowcolor{TableAccent}
CD-1M full (Ours)
& \textbf{1.53} / \textbf{8.46} & \textbf{0.27} & \textbf{0.21} & \textbf{3.22} & \textbf{0.19} & \textbf{0.30} & \textbf{0.69} & \textbf{0.6985} \\
\midrule
\multicolumn{9}{@{}l}{\textit{Effect of training strategy}} \\
Direct T2AV
& 1.21 / 9.05 & 0.20 & 0.16 & 2.98 & 0.13 & 0.25 & 0.64 & 0.5630 \\
\rowcolor{TableAccent}
Full DARC (Ours)
& \textbf{1.53} / \textbf{8.46} & \textbf{0.27} & \textbf{0.21} & \textbf{3.22} & \textbf{0.19} & \textbf{0.30} & \textbf{0.69} & \textbf{0.6985} \\
\bottomrule
\end{tabular}
\end{table*}

\noindent\textbf{Quantitative Comparison.}
\textbf{\textit{1)}} For basic video and audio quality, \OurModel achieves the best or competitive scores across most metrics.
These low-level quality metrics are closely related to the capability of the underlying backbone, and \OurModel remains competitive with, or even surpasses, pipelines that generate shots separately.
Compared with the base LTX-2.3, \OurModel obtains clear improvements, indicating that long-form adaptation on CineDance-1M does not degrade the pretrained visual-audio prior and further demonstrates the effectiveness of our training strategy and the high quality of the dataset.
\textbf{\textit{2)}} For audio-video synchronization and prompt alignment, \OurModel also achieves the best or competitive performance.
The relatively low Sync-C scores across methods are mainly caused by challenging benchmark cases with off-screen speech, large facial motion, or non-frontal faces, where lip-centric synchronization models become less reliable.
Nevertheless, \OurModel improves both shot-level and video-level prompt following, showing that structured CineDance training helps the model better understand multi-shot narrative conditions.
\textbf{\textit{3)}} For narrative continuity, \OurModel obtains stronger identity and scene consistency than existing baselines, demonstrating its ability to preserve recurring characters and environments across shot transitions, which is consistent with the qualitative observations in Fig.~\ref{fig:compare}.
Against the base LTX-2.3, the improvements further confirm that the gains come from CineDance adaptation rather than the pretrained backbone alone.

\noindent\textbf{Shot-structure response analysis.}
For fair comparison, we report Shot Structure Response (SSR) separately in Tab.~\ref{tab:ssr}, since Paradigm-2 baselines receive explicit ground-truth shot-layout conditions.
Paradigm-1  methods obtain identical SSR scores because their normalized shot partitions are the same after stitching.
Their $S_{\mathrm{cnt}}$ reaches 1.0 because these methods are given the target number of shots and generate each shot separately, indicating that they are not fully free from shot-structure information either.
However, their lower $S_{\mathrm{seg}}$ shows that matching the shot count alone does not guarantee a faithful temporal layout, as uniform or suboptimal average shot durations may still deviate from the ground-truth shot distribution.
The base LTX-2.3 obtains substantially weaker SSR, confirming that the pretrained backbone has limited response to shot-level structure, consistent with our observations in Sec.~\ref{sec:intro} and Sec.~\ref{sec:method_scaffold}.
Although \OurModel does not receive explicit shot-count or shot-layout control, its $S_{\mathrm{cnt}}$ remains close to 1.0, and its $S_{\mathrm{seg}}$ and final SSR surpass Paradigm-1 methods, demonstrating stronger intrinsic shot-structure response and validating SSR as a meaningful and robust metric for shot-structure benchmarking.

\subsection{Ablation Study}

\noindent\textbf{Ablation on datasets.}
We first ablate the role of training data while keeping the backbone, training strategy,  and optimization budget unchanged.
Specifically, we compare five variants:
\textbf{\textit{1)}} training on 1M randomly sampled clips from OpenHumanVid~\cite{openhumanvid},
\textbf{\textit{2)}} training on the full OpenHumanVid,
\textbf{\textit{3)}} training on isolated shot-only clips from \Ours,
\textbf{\textit{4)}} training on \Ours without structured shot-level annotations,
and \textbf{\textit{5)}} training on the full \Ours.
OpenHumanVid~\cite{openhumanvid} serves as an external human-centric audio-video baseline.
The shot-only variant uses the shot-level annotation of each isolated shot as its prompt, without modeling multi-shot context.
As shown in Tab.~\ref{tab:ablation}, short-clip training mainly preserves the basic capability of the LTX-2.3 backbone, but brings limited improvement on narrative-oriented metrics such as multi-shot consistency and SSR.
Training on \Ours without structured shot-level annotations yields moderate gains on these narrative metrics, suggesting that multi-shot videos themselves provide useful supervision for multi-shot long-form generation.
However, this variant remains clearly below the full setting, showing that structured shot-level annotations are necessary for learning fine-grained prompt following and controllable shot organization.
These results demonstrate the value of CineDance-1M from two complementary aspects: \textbf{\textit{1)}} long-form multi-shot video data and \textbf{\textit{2)}} structured shot-wise annotations.

\noindent\textbf{Ablation on methods.}
We further ablate the training strategy under the same \Ours data setting.
The Direct T2AV variant uses the same structured textual prompts as Full DARC, but directly optimizes the text-only audio-video generation objective without any reference-scaffolded training.
Full DARC instead starts from reference-scaffolded multi-shot learning and progressively removes the visual and temporal scaffolds until the model is trained under the same text-only condition used at inference.
As shown in Tab.~\ref{tab:ablation}, Direct T2AV preserves the basic generation ability of the backbone, but yields limited gains in shot-level prompt following, cross-shot continuity, and shot-structure response.
Full DARC consistently improves all evaluated dimensions, indicating that progressive scaffold removal provides a more effective easy-to-hard training path for learning multi-shot organization from \Ours.

\begin{table}[t]
    \centering
    \caption{
Human Evaluation Results. We report the Mean Opinion Score (1-5 scale, higher is better) across the six proposed dimensions. Best results are \textbf{bolded}.
    }
    \label{tab:human-study-mos}
    \setlength{\tabcolsep}{3.5pt}
    \resizebox{\linewidth}{!}{
        \renewcommand{\arraystretch}{1.15}
        \begin{tabular}{@{}l|cccccc@{}}
            \toprule
            \textbf{Method} & \textbf{VQ} & \textbf{AQ} & \textbf{Sync} & \textbf{Align.} & \textbf{Cont.} & \textbf{SSR} \\
            \midrule
            CineTrans~\cite{cinetrans} & 2.92 & -- & -- & 3.08 & 2.71 & 3.15 \\
            Mask2DiT~\cite{mask2dit} & 3.91 & -- & -- & 2.72 & 2.85 & 3.18 \\
            MultiShotMaster~\cite{multishotmaster} & 3.32 & -- & -- & 3.12 & 2.31 & 3.62 \\
            HoloCine~\cite{holocine} & 3.18 & -- & -- & 3.75 & 2.58 & 3.61 \\
            % \midrule
            StoryMem~\cite{storymem} & 3.72 & -- & -- & 2.85 & 3.31 & 3.42 \\
            VGoT~\cite{vgot} \Omark & 3.42 & -- & -- & 3.55 & 3.35 & 3.41 \\
            VGoT~\cite{vgot} \Lmark & 3.48 & 3.72 & 3.65 & 3.41 & 3.22 & 3.31 \\
            MovieAgent~\cite{movieagent} \Omark & 3.55 & -- & -- & 3.51 & 3.21 & 3.58 \\
            MovieAgent~\cite{movieagent} \Lmark & 3.61 & 3.85 & 3.71 & 3.54 & 3.25 & 3.45 \\
            STAGE~\cite{stage} \Omark & 3.28 & -- & -- & 3.68 & 2.52 & 3.58 \\
            STAGE~\cite{stage} \Lmark & 3.32 & 3.61 & 3.41 & 3.38 & 2.81 & 3.42 \\
            LTX-2.3 (Base)~\cite{ltx-2} & 3.75 & \textbf{4.05} & 3.12 & 3.32 & 2.88 & 2.61 \\
            \midrule
            \rowcolor{TableAccent}
            \textbf{CineDance (Ours)} & \textbf{4.12} & 3.98 & \textbf{4.11} & \textbf{4.18} & \textbf{4.25} & \textbf{4.38} \\
            \bottomrule
        \end{tabular}
    }
\end{table}

\subsection{Human Study}
\noindent\textbf{Human preference study.}
As shown in Tab.~\ref{tab:human-study-mos}, \OurModel achieves the best Mean Opinion Score (MOS) on five out of six dimensions, including video quality, audio-video synchronization, prompt alignment, narrative continuity, and shot-structure response, while remaining competitive in audio quality.
Compared with the base LTX-2.3 model, \OurModel shows substantial improvements on structure-related dimensions, demonstrating that CineDance adaptation effectively enhances multi-shot controllability, cross-shot coherence, and narrative organization.
Compared with existing multi-shot baselines, the gains are especially pronounced in continuity and SSR, indicating that human evaluators perceive more stable recurring characters, more coherent scenes, and clearer shot transitions.
These results show that CineDance improves not only automatic scores but also human-perceived multi-shot generation quality.
% \noindent\textbf{Human study.} 
% To validate our automated metrics, 10 independent evaluators assessed a sampled test subset under randomized, double-blind conditions. Using a 5-point Likert scale, evaluators rated videos across our six core dimensions: \textbf{\textit{1) Video Quality}} (fidelity \& smoothness), \textbf{\textit{2) Audio Quality}} (clarity), \textbf{\textit{3) AV Sync}} (lip-sync \& alignment), \textbf{\textit{4) Prompt Alignment}} (faithfulness), and \textbf{\textit{5) Narrative Continuity}} (identity \& scene persistence). To strictly prevent subjective bias, all source models were anonymized, and the presentation order was completely randomized. Crucially, these subjective ratings achieve a strong Spearman rank correlation ($\rho\!=\!0.92$, $p\!<\!10^{-4}$) with our automated leaderboard across the 14 evaluated systems, firmly validating the human-aligned design of \OurBench. Furthermore, \OurModel ranks first on every dimension (Video Q.\ \textbf{4.02}, Audio Q.\ \textbf{3.83}, AV Sync \textbf{3.71}, Prompt Align.\ \textbf{3.92}, Narr.\ Cont.\ \textbf{4.05}), confirming its comprehensive superiority. 
% \textit{Detailed human preference results and human-alignment analysis are provided in Appendix~\ref{app:human-study}.}

\begin{figure}[t]
    \centering
    \includegraphics[width=\linewidth]{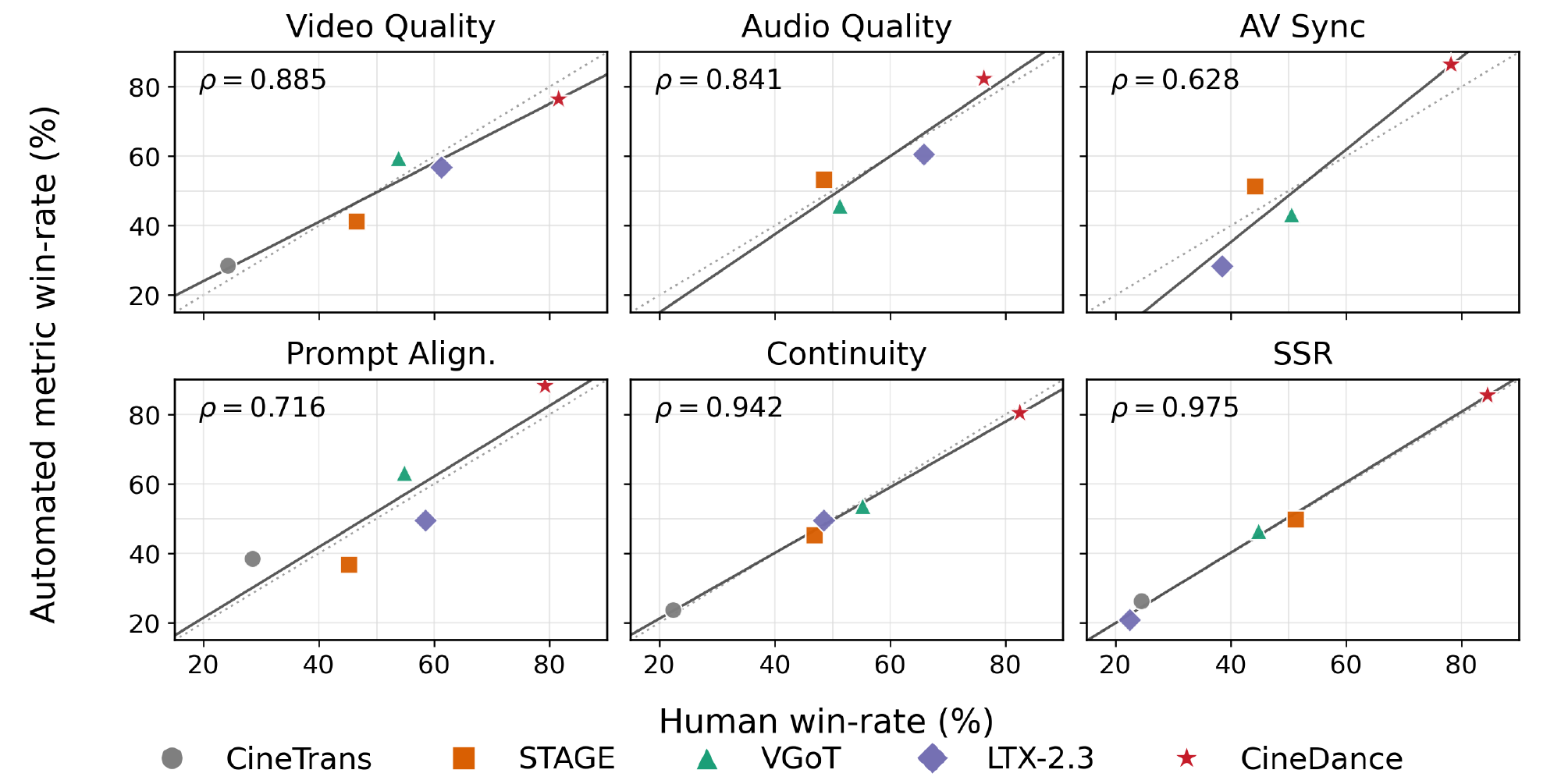}
\caption{
Human alignment of the CineBench automatic metrics.
Each subplot compares model-level human win rates with the corresponding automated metric win rates across six evaluation dimensions.
The reported Spearman rank correlation $\rho$ indicates strong human alignment for most dimensions.
}

\label{fig:human_alignment}
\end{figure}

\noindent\textbf{Human Alignment Analysis.}
Fig.~\ref{fig:human_alignment} further examines whether CineBench automatic metrics are aligned with human judgments.
For each evaluation dimension, we compute the model-level win rate induced by automatic scores and compare it with the corresponding human win rate.
The two rankings show positive correlations across all six dimensions, indicating that the proposed automatic suite generally reflects human preferences.
In particular, Continuity and SSR achieve the strongest correlations, with Spearman coefficients of $\rho=0.942$ and $\rho=0.975$, respectively, suggesting that our annotation-grounded continuity metrics and shot-structure response metric are well aligned with human perception of multi-shot coherence.
Video Quality and Audio Quality also show high correlations, while AV Sync exhibits a relatively weaker but still positive correlation.
Overall, the alignment analysis supports CineBench as a meaningful automatic evaluation suite for multi-shot, narrative-driven audio-video generation.

\section{Conclusion}
\label{sec6}
In this paper, we introduced \Ours, a pioneering million-scale 1080p dataset designed to overcome the critical bottleneck in multi-shot, long-form joint audio-video generation. Driven by film-theory-inspired narrative parsing and hierarchical dual-modal annotation, our curation pipeline guarantees exceptional semantic alignment. Alongside the dataset, we established \OurBench for human-aligned evaluation and developed a robust baseline, \OurModel, which demonstrates precise cross-modal synchronization and spatio-temporal consistency, successfully bridging the divide between isolated short clips and cohesive narrative synthesis. \\
\noindent\textbf{Limitations, broader impacts and ethical considerations.} While \OurBench is currently bounded at 1080p resolution, we intend to advance ultra-high-definition applications in future work by releasing an additional 4K multi-shot audio-visual dataset. Furthermore, we acknowledge that highly realistic synthesis inherently exacerbates deepfake risks, underscoring the urgent need for robust deepfake detection frameworks to safeguard information authenticity.
% Finally, to comprehensively address ethical concerns, we first ensure all source data is strictly copyright-compliant, and furthermore release \Ours under a 
% % \emph{metadata-only}, \emph{gated}, and \emph{takedown-supported} scheme 
% gated and takedown-supported scheme 
% % (detailed in Appendix.~\ref{app:ethics-and-release}) 
% to avoid redistributing raw bitstreams.

\section*{Declarations}

\noindent\textbf{Data Availability.}
CineBench prompts, evaluation metadata, and metric implementations will be made publicly available on the project page upon acceptance or publication.
For CineDance-1M, we will publicly release the structured annotations and metadata associated with the dataset, together with filtering scripts and reconstruction instructions.
% where permitted by the original data licenses and source-platform restrictions.
For video sources derived from public platforms such as YouTube, we will provide metadata-based access rather than redistributing raw video files.
For self-collected sources for which redistribution is permitted, access to the corresponding data will be provided through a gated application process for research use.
Raw source videos subject to third-party copyright, platform terms, or redistribution restrictions will not be directly redistributed.
We will maintain a takedown-supported access protocol and promptly remove or restrict access to any data upon valid copyright, privacy, or source-platform requests. \\
\noindent\textbf{Code Availability.}
The code for data processing, benchmark construction, evaluation, and model adaptation will be released on the project page upon acceptance or publication.

% \appendix

% \section{Appendix Placeholder}\label{secA1}

% Appendix content will be added here. See the supplementary material for pipeline validation details (Appendices \ref{app:pipeline-data-format}, \ref{app:pipeline-narrative-parsing}, \ref{app:pipeline-annotation}), benchmark details (\ref{app:pipeline-cinebench}), metric implementation (\ref{sec:suppl:cinebench-metric-impl}), and human study results (\ref{app:human-study}).

\bibliographystyle{spmpsci}
\bibliography{main}

\end{sloppypar}
\end{document}